%% file: main.tex
\documentclass{article}
\usepackage{microtype}
\usepackage{graphicx}
\usepackage{subfigure}
\usepackage{booktabs} 
\usepackage{hyperref}
\usepackage[accepted]{icml2024} 
\usepackage{amsmath}
\usepackage{amssymb}
\usepackage{mathtools}
\usepackage{amsthm}
\usepackage[capitalize,noabbrev]{cleveref}
\usepackage[american]{babel}
\usepackage{natbib}
\usepackage{booktabs}
\usepackage[utf8]{inputenc}
\usepackage{times}
\usepackage{bbold}
\usepackage{graphicx}
\usepackage{rotating}
\usepackage[algo2e]{algorithm2e}
\usepackage{pifont}
\usepackage{hyperref}
\usepackage{makecell}
\usepackage{capt-of}
\usepackage{tcolorbox}
\usepackage{bbm}
\usepackage{multirow}
\usepackage[textsize=small,disable]{todonotes} 

\icmltitlerunning{Matroid Semi-Bandits in Sublinear Time}
\icmlsetsymbol{equal}{*}
\input{marcos}

\begin{document}
\twocolumn[
\icmltitle{Matroid Semi-Bandits in Sublinear Time}
\begin{icmlauthorlist}
\icmlsetsymbol{intern}{\S}
\icmlauthor{Ruo-Chun Tzeng}{kth,intern}
\icmlauthor{Naoto Ohsaka}{ca}
\icmlauthor{Kaito Ariu}{ca}
\end{icmlauthorlist}
\icmlaffiliation{kth}{EECS, KTH Royal Institue of Technology, Sweden}
\icmlaffiliation{ca}{AI Lab, CyberAgent, Japan}
\icmlcorrespondingauthor{Ruo-Chun Tzeng}{rubys88684@gmail.com}
\icmlkeywords{combinatorial semi-bandits}
\vskip 0.3in
]
\printAffiliationsAndNotice{\S Work done during an internship at CyberAgent.} 
\begin{abstract}
We study the matroid semi-bandits problem, where at each round the learner plays a subset of $K$ arms from a feasible set, and the goal is to maximize the expected cumulative linear rewards. 
Existing algorithms have per-round time complexity at least $\Omega(K)$, which becomes expensive when $K$ is large. 
To address this computational issue, we propose {\tt FasterCUCB} whose sampling rule takes time sublinear in $K$ for common classes of matroids: $\bigO(D\,\polylog{K}\,\polylog{T})$ for uniform matroids, partition matroids, and graphical matroids, and $\bigO(D\sqrt{K}\polylog{T})$ for transversal matroids. 
Here, $D$ is the maximum number of elements in any feasible subset of arms, and $T$ is the horizon.
Our technique is based on dynamic maintenance of an approximate maximum-weight basis over inner-product weights. Although the introduction of an approximate maximum-weight basis presents a challenge in regret analysis, we can still guarantee an upper bound on regret as tight as {\tt CUCB} in the sense that it matches the gap-dependent lower bound by \citet{kveton2014matroid} asymptotically.

\end{abstract}

\input{sections/1.intro}
\input{sections/2.prelim}
\input{sections/3.related}
\input{sections/4.subroutine}
\input{sections/5.regret}
\input{sections/6.conclusion}
\bibliographystyle{plainnat}
\bibliography{ref}
\onecolumn
\appendix
\newpage
\input{app_notation}
\newpage
\input{app_related}
\newpage
\input{app_membership}
\newpage
\input{app_matroids}
\newpage
\end{document}

%% file: marcos.tex
\DeclareMathOperator*{\argmax}{argmax} 

\theoremstyle{plain}
\newtheorem{theorem}{Theorem}[section]

\newtheorem{lemma}[theorem]{Lemma}
\newtheorem{corollary}[theorem]{Corollary}
\theoremstyle{definition}

\newtheorem{assumption}[theorem]{Assumption}
\theoremstyle{remark}
\newtheorem{remark}[theorem]{Remark}

\newenvironment{proofof}[1]{\paragraph{Proof of #1:}}{\hfill$\square$}
\newenvironment{pf}{\paragraph{Proof:}}{\hfill$\square$}

\makeatletter
\newtheorem*{rep@theorem}{\rep@title}
\newcommand{\newreptheorem}[2]{%
	\newenvironment{rep#1}[1]{%
		\def\rep@title{#2 \ref{##1}}%
		\begin{rep@theorem}}%
		{\end{rep@theorem}}}
\makeatother
\newreptheorem{theorem}{Theorem}
\newreptheorem{lemma}{Lemma}
\newreptheorem{proposition}{Proposition}
\newreptheorem{corollary}{Corollary}
\newreptheorem{claim}{Claim}
\newreptheorem{assumption}{Assumption}

\newcommand{\reals}{\ensuremath{\mathbb{R}}\xspace}
\newcommand{\ints}{\ensuremath{\mathbb{N}}\xspace}
\newcommand{\expectation}[2]{\ensuremath{\mathbb{E}_{#1}\!\left[#2\right]}\xspace}

\newcommand{\prob}[2]{\ensuremath{\mathbb{P}_{#1}\!\left[#2\right]}\xspace}

\newcommand{\bmu}{\ensuremath{\bd{\mu}}\xspace}

\newcommand{\ba}{\ensuremath{\bd{a}}\xspace}

\newcommand{\bdf}{\ensuremath{\bd{f}}\xspace}

\newcommand{\bx}{\ensuremath{\bd{x}}\xspace}
\newcommand{\by}{\ensuremath{\bd{y}}\xspace}

\newcommand{\bv}{\ensuremath{\bd{v}}\xspace}

\newcommand{\be}{\ensuremath{\bd{e}}\xspace}
\newcommand{\bh}{\ensuremath{\bd{h}}\xspace}

\newcommand{\bq}{\ensuremath{\bd{q}}\xspace}

\newcommand{\gap}[1]{\ensuremath{\triangle_{#1}}\xspace}
\newcommand{\supp}[1]{\ensuremath{\text{ supp}\left(#1\right)}\xspace}

\newcommand{\norm}[1]{\ensuremath{\left\|{#1}\right\|}\xspace}

\newcommand{\abs}[1]{\ensuremath{\lvert{#1}\rvert}\xspace}

\newcommand{\veczeros}[1]{\ensuremath{\mathbf{0}_{#1}}\xspace}

\newcommand{\inner}[2]{\ensuremath{\left\langle{#1},{#2}\right\rangle}\xspace}

\newcommand{\ind}[1]{\ensuremath{\mathbbm{1}\!\left\{{#1}\right\}}\xspace}
\newcommand{\poly}[1]{\ensuremath{\text{ poly}\left({#1}\right)}\xspace}
\newcommand{\polylog}[1]{\ensuremath{\text{polylog}\left({#1}\right)}\xspace}
\newcommand{\bin}[1]{\ensuremath{\text{BIN}_{#1}}\xspace}
\newcommand{\ub}{\ensuremath{\text{ub}}\xspace}
\newcommand{\lb}{\ensuremath{\text{lb}}\xspace}  
\newcommand{\err}{\ensuremath{\eta}\xspace}  

\newcommand{\bd}[1]{\boldsymbol{#1}\xspace}
\newcommand{\decls}{\ensuremath{\mathcal{X}}\xspace}
\newcommand{\icls}{\ensuremath{\mathcal{I}}\xspace}
\newcommand{\matcls}{\ensuremath{\mathcal{M}}\xspace}

\newcommand{\istar}[1]{\ensuremath{\bd{i}^\star\!\left(#1\right)}\xspace}
\newcommand{\ist}{\bd{i}^\star}

\SetCommentSty{commfont}

\newcommand{\bigO}{\ensuremath{\mathcal{O}}\xspace}

\newcommand{\dom}{\ensuremath{\mathrm{dom}}\xspace}

\newcommand{\kaitocomment}[1]{\todo[inline]{Kaito: #1}}

%% file: sections/1.intro.tex
\section{Introduction}
Matroid semi-bandits model many real-world tasks.
An instance of matroid semi-bandit is described by $([K],\decls,\bmu)$, where $[K]\triangleq\{1,\cdots,K\}$ is the ground set, each $k\in[K]$ is associated with a probability distribution $\nu_k$ with mean $\mu_k$, and $\decls\subseteq\{0,1\}^K$ is the set of bases of a given matroid $\matcls=([K],\icls)$ of rank $D$.
At each round $t\in[T]$, the learner pulls an action $\bx(t)\in\decls$ and observes a {\it semi-bandit} feedback, i.e., $y_k(t)\sim\nu_k$ iff $x_k(t)=1$.
This formulation can be used to model online advertisting and news selection \cite{kale2010non} with $\matcls$ as a uniform matroid. Ad placement \cite{bubeck2013multiple,streeter2009online} and diversified recommendation \cite{abbassi2013diversity} can be modeled with $\matcls$ as a partition matroid. Network routing \cite{kveton2014matroid} can be modeled with $\matcls$ as a graphical matroid.
Task assignment \cite{chen2016pure} can be modeled with $\matcls$ as a transversal matroid. 

Popular algorithms include Combinatorial Upper Confidence Bound ({\tt CUCB}) \cite{gai2012combinatorial,chen2013combinatorial,kveton2014matroid,kveton2015tight}, Combinatorial Thompson Sampling ({\tt CTS}) \cite{wang2018thompson,kong2021hardness,perrault2022combinatorial}, and the instance-specifically optimal algorithm KL-based Efficient Sampling for Matroids ({\tt KL-OSM}) \cite{talebi2016optimal}. 
All of these algorithms rely on a greedy algorithm (see Algorithm~\ref{alg:greedy}) to determine the action to be pulled. The greedy algorithm takes time at least $\Omega(K)$ and at most $\bigO(K(\log K+\mathcal{T}_{\mathrm{member}}))$, where $\mathcal{T}_{\mathrm{member}}$ is the time taken to determine whether $\bx+\be_k\in\icls$ for some $(\bx,k)\in\icls\times[K]$, and $\be_k$ is the $k$-th canonical unit
vector. 
However, when the number $K$ of arms is large, performing the greedy algorithm at each round can become expensive. There is a need to develop a matroid semi-bandit algorithm with per-round time complexity sublinear in $K$.

In this work, we present {\tt FasterCUCB} (Algorithm~\ref{alg:faster-cucb}), the first sublinear-time algorithm for matroid semi-bandit. 
The design of {\tt FasterCUCB} is based on {\tt CUCB}, but with a much faster sampling rule which takes time sublinear in $K$ for many classes of matroids.
For uniform matroids, partition matroids, and graphical matroids, it has per-round time complexity of $\bigO(D\,\polylog{K}\polylog{T})$, which is optimal up to a polylogarithmic factor as compared to the trivial lower bound of $\Omega(D)$. 
For transversal matroids, the per-round time complexity is $\bigO(D\sqrt{K}\,\polylog{T})$, which is still sublinear in $K$ when $D = \bigO(K^{\frac{1}{2}-\epsilon})$ for any $\epsilon > 0$.
{\tt FasterCUCB} trades the accuracy for computational efficiency. In other words, the action computed by the sampling rule of {\tt FasterCUCB} is an {\it approximation} to the optimal solution computed by the sampling rule of {\tt CUCB}. This introduces difficulty in the regret analysis because prior analysis of {\tt CUCB} \cite{kveton2014matroid} requires the exact solution.
What is interesting is that we can still guarantee the same regret upper bound asymptotically as prior analysis of {\tt CUCB}.

\input{sections/tab.per-round-time}

To develop a sublinear-time sampling rule,
we present a dynamic algorithm for maintaining maximum-weight base \emph{over inner product weights} (\cref{sec:dynamic}). 
There have been many sublinear-time algorithms for dynamic maximum-weight base maintenance
(see \cref{sec:related-works}),
which, however, may not be directly used in {\tt FasterCUCB} because
\emph{all} arm weights representing the UCB index can change simultaneously at each round.
Our insight for addressing this issue is that
the UCB index of each arm $k$ at round $t$
can be decomposed into an inner product of the following two-dimensional vectors:
(1) a \emph{feature},
which depends on $k$ and is supposedly
a pair of the empirical reward estimate and radius of confidence interval, and
(2) a \emph{query}, 
which depends only on round $t$.
Our proposed dynamic algorithm consists of two speeding-up techniques.
One is \emph{feature rounding},
which rounds each feature
into a few bins so as to reduce the number of distinct features to consider.
The other is the \emph{minimum hitting set} technique,
which allows us to compute a small number of queries in advance and
correctly identify an (approximate) maximum-weight base for \emph{any} query.

Sections are organized as follows. We introduce matroid semi-bandits and basic concepts in Section~\ref{sec:pre}. We review relevant literature in  Section~\ref{sec:related-works}. 
We develop a dynamic algorithm for maintaining a maximum-weight base over inner product weights in Section~\ref{sec:dynamic}.
We propose {\tt FasterCUCB} based on the algorithms developed in Section~\ref{sec:dynamic} and analyzed its regret and time complexity in Section~\ref{sec:regret-analysis}.

%% file: sections/tab.per-round-time.tex
\begin{table*}[!ht]
    \centering
    \begin{tabular}{c|c|c}\toprule
     & {\tt CUCB} & {\tt FasterCUCB} \\\toprule
    Per-round Time Complexity 
        & $\bigO(K(\log K +\mathcal{T}_{\mathrm{member}}))$
        & $\bigO(D\,\polylog{ T}\mathcal{T}_{\mathrm{update}}(\mathcal{A}))$\\\toprule
    Uniform Matroid 
        & $\bigO(K\log K)$
        & $\bigO(D\log K\,\polylog{T})$\\
    Partition Matroid 
        & $\bigO(K\log K)$ 
        & $\bigO(D\log K\,\polylog{T})$\\
    Graphical Matroid 
        & $\bigO(K\log K)$
        & $\bigO(D\,\polylog{K}\,\polylog{T})$\\
    Transversal Matroid 
        & $\bigO(K(\log K+DK))$
        & $\bigO(D\sqrt{K}\,\polylog{T})$\\
    \bottomrule
    \end{tabular}
    \caption{
        Per-round time complexity of {\tt CUCB} \cite{kveton2014matroid} and {\tt FasterCUCB} (Algorithm~\ref{alg:faster-cucb}) for different classes of matroids.
        $K$ is the number of arms and $D$ is the maximum number of elements in any action in $\decls$.
        $\mathcal{T}_{\mathrm{member}}$ for different matroids is discussed in Appendix~\ref{sec:membership}.
        $\mathcal{T}_{\mathrm{update}}(\mathcal{A})$ for different matroids is discussed in Section~\ref{sec:related-works}.
    }
    \label{tab:per-round-time}
\end{table*}

%% file: sections/2.prelim.tex
\section{Preliminaries}
\label{sec:pre}

We use $[n]$ to denote the set $\{1,\cdots,n\}$.
We use $\istar{\bmu}$ to denote any element in $\argmax_{\bx\in\decls}\inner{\bmu}{\bx}$, and when it is clear from the context, we drop $\bmu$ from $\istar{\bmu}$ and write $\ist$.
We use $\supp{\cdot}$ to denote the support set of a given vector.
We use $\be_k$ to denote the vector with $1$ only on the $k$-th row and $0$'s elsewhere, and use $\veczeros{K}$ to denote a $K$-dimensional vector with $0$ on every row.
We use $\log$ with base $e$.
See Appendix~\ref{app:notation} for a table of notation.

\paragraph{Matroid.}
A matroid is described by $\matcls\triangleq([K],\icls)$, where $[K]$ is called the {\it ground set} and $\icls\subseteq\{0,1\}^K$ is the set of {\it independent sets} satisfying (i) {\it hereditary property}, i.e., if $\supp{\by}\subset\supp{\bx}$ and $\bx\in\icls$, then $\by\in\icls$; and (ii) {\it augmentation property}, i.e.,  if $\bx,\by\in\icls$ and $\supp{\by}\subset\supp{\bx}$, then there exists $j\in\supp{\bx}\backslash\supp{\by}$ such that $\by+\be_j\in\icls$.
We said $\bx\in\icls$ is a basis if $\supp{\bx}$ is {\it maximal}, i.e., there does not exist $\by\in\icls$ such that $\supp{\bx}\subset\supp{\by}$. 
All bases have the same cardinality, which is called the {\it rank} of the matroid. For $\bv\in\reals_+^K$, a maximum-weight basis $\istar{\bv}\in\argmax_{\bx\in\decls}\sum_{k=1}^Kv_kx_k$ can be found by a greedy algorithm (Algorithm~\ref{alg:greedy}) in $\bigO(K(\log K+\mathcal{T}_{\mathrm{member}}))$ time, where $\mathcal{T}_{\mathrm{member}}$ is the time taken for the membership oracle to determine whether $\bx+\be_k\in\icls$ for some $\bx\in\icls$ and some $k\in[K]\backslash\supp{\bx}$.

\begin{algorithm}[h!]
\caption{A greedy maximum-weight basis algorithm}
\label{alg:greedy}
\footnotesize
    \KwIn{$\bv\in\reals^K$ and the bases $\decls\subseteq\{0,1\}^K$.}
    Sort $\bv$ in non-increasing order: $v_{\gamma(1)}\geq\cdots\geq v_{\gamma(K)}$;
    
    $\bx=\veczeros{K}$; $i=1$;
    
    \While{$\norm{\bx}_0<D$}{
        \uIf{$\bx+\be_{\gamma(i)}\in\icls$}{
            $\bx\leftarrow\bx+\be_{\gamma(i)}$
        }
        $i\leftarrow i+1$;
    }
\end{algorithm}


\paragraph{Matroid semi-bandits.} An instance of matroid semi-bandit is described by $([K],\decls,\bmu)$, where $[K]$ is the ground set, $\decls\subseteq\{0,1\}^K$ is the set of bases of the given matroid $\matcls\triangleq([K],\icls)$ of rank $D$, 
and $\icls\subseteq\{0,1\}^K$ is the set of independent sets.
Each $k\in[K]$ is associated with a distribution $\nu_k$ with mean $\mu_k$.
The learner knows the matroid $\matcls$, and aims to learn the best action $\istar{\bmu}\in\argmax_{\bx\in\decls}\inner{\bmu}{\bx}$ by playing a game with the environment: At each round $t\in\ints$, the learner plays an action $\bx(t)$, and the environment draws a noisy reward $y_k(t)\sim\nu_k$ for each arm $k\in[K]$ and reveals $y_k(t)$ to the learner iff $k\in\supp{\bx(t)}$.
We assume arms' rewards are bounded: 
\begin{assumption}\label{ass:lm}
    Assume the support of each arm $\nu_k$ is a subset of $[a,b]$ and $0<a<b<\infty$.
\end{assumption}
The performance is measured by expected regret:
\[
R(T)\triangleq T\inner{\bmu}{\istar{\bmu}}-\expectation{}{\sum_{t=1}^T\inner{\bmu}{\bx(t)}},
\]
which is the difference between the expected cumulative reward of the learner and that of an algorithm who knows $\bmu$ and always selects the best action $\istar{\bmu}$.

\paragraph{Common classes of matroids.}
Refer to Chapter 39  \cite{schrijver2003combinatorial} or Chapter 1  \cite{oxley2011matroid} for more details.
\begin{itemize}
    \item A {\it uniform matroid} $([K],\icls)$ of rank $D$ has independent sets $\icls=\{S\subseteq[K]:\abs{S}\leq D\}$ and the bases $\decls$ consist of subsets whose cardinalities are exactly $D$.
    \item A {\it partition matroid} $([K],\icls)$ of rank $D$ is given a partition $S_1,\cdots,S_D$ of the ground set $[K]$, the independent sets $\icls=\left\{S:\abs{S\cap S_i}\leq 1,\forall i\in[D]\right\}$, and the bases $\decls$ are subsets that choose exactly one element from each of the $D$ sets.
    \item A {\it graphical matroid} is given a graph $G=(V,E)$ with $K$ edges, the bases $\decls$ consist of all spanning forests in $G$, and the rank $D$ is $\abs{V}$ minus the number of connected components in $G$. 
    \item A {\it transversal matroid} is given a bipartite graph $G=([K]\cup V,E)$ with a bipartition $([K],V)$, $\abs{V}\leq K$, the independent sets $\icls$ consist of $S\subseteq [K]$ such that there is a matching of $S$ to $\abs{S}$ vertices in $V$, and $\decls$ is the set of endpoints in $[K]$ of all maximum matchings in $G$. 
\end{itemize}
We discuss the query time $\mathcal{T}_{\mathrm{member}}$ of membership oracle in Appendix~\ref{sec:membership}.
For more examples on semi-bandits under different matroid constraints, we refer the readers to \cite{kveton2014matroid} for linear matroids, and \cite{kveton2014learning} for polymatroid semi-bandits. 

\paragraph{CUCB.}
The action selected by {\tt CUCB} \cite{gai2012combinatorial,chen2013combinatorial,kveton2014learning} at round $t\in\ints$ is: 
\begin{equation}\label{eq:combucb}
\bx(t)\in\argmax_{\bx\in\decls}\sum_{k=1}^K\left(\hat{\mu}_k(t-1)+\frac{\lambda_t}{\sqrt{N_k(t-1)}}\right)x_k,
\end{equation}
where $\hat{\mu}_k(t)\triangleq\frac{1}{N_k(t)}\sum_{s=1}^ty_k(s)\ind{x_k(s)=1}$ is the empirical reward estimate, $N_k(t)\triangleq\sum_{s=1}^t\ind{x_k(s)=1}$ is the number of pulls of arm $k$, and $\lambda_t>0$ controls the confidence width.
The value $\hat{\mu}_k(t-1)+\frac{\lambda_t}{\sqrt{N_k(t-1)}}$ is called the {\it UCB index} of arm $k$.
In \cite{kveton2014matroid}, Eq.~\eqref{eq:combucb} is solved by a $\bigO(K(\log K+\mathcal{T}_{\mathrm{member}}))$-time greedy algorithm shown in Algorithm~\ref{alg:greedy}.
In Section~\ref{sec:dynamic}, we will develop a faster algorithm for solving  Eq.~\eqref{eq:combucb} with the following reformulation:
\[
\bx(t)\in\argmax_{\bx\in\decls}
\sum_{k=1}^K\inner{\bdf_k}{\bq}x_k,
\]
where $\bdf_k=(\hat{\mu}_k(t-1),\frac{1}{\sqrt{N_k(t-1)}})$ and $\bq=(1,\lambda_t)$.

%% file: sections/3.related.tex
\section{Related Works}\label{sec:related-works}
\paragraph{Semi-bandits and sublinear-time bandits.}
We provide an extensive survey on related bandit literature in Appendix~\ref{sec:further-related-works}. To summarize here, for semi-bandit algorithms, {\tt CUCB} \cite{gai2012combinatorial,chen2013combinatorial,kveton2014matroid}, {\tt CTS} \cite{wang2018thompson,kong2021hardness,perrault2022combinatorial} and {\tt KL-OSM} \cite{talebi2016optimal} all rely on a $\bigO(K(\log K+\mathcal{T}_{\mathrm{member}}))$-time greedy algorithm to compute the action to be pulled. In contrast, our {\tt FasterCUCB}, as far as we know, is the first semi-bandit algorithm having per-round time complexity of $o(K)$. For linear bandits, there exist several works \cite{jun2017scalable,yang2022linear} on reducing the per-round complexity to be sublinear in the number of arms. But, their results transferred to our setting are worse than what we have obtained both in terms of regret bound and the time complexity (see the discussion in Appendix~\ref{sec:further-related-works}).

\paragraph{Dynamic maintenance of maximum-weight base of a matroid}
Here, we review existing dynamic algorithms for maintaining a maximum-weight base of a matroid.
In a standard (fully-)dynamic setting,
we are given a weighted matroid $\matcls = ([K], \icls)$, where
each arm's weight dynamically changes over time in an online manner.
The objective is to maintain any (exact or approximate) maximum-weight basis of $\matcls$ over up-to-date arm weights as efficiently as possible.
We use $\mathcal{T}_{\mathrm{update}}(\mathcal{A})$ to denote
the time complexity of a dynamic algorithm $\mathcal{A}$
required for updating an (approximate) maximum-weight base according to
the change of a single arm weight.
The best-known bound of
$\mathcal{T}_{\mathrm{update}}(\mathcal{A})$
for each matroid class is summarized as follows:
For graphic matroids,
a maximum-weight basis can be updated
in $\bigO(\sqrt{K})$ worst-case time \cite{frederickson1985data,eppstein1997sparsification} and 
in $\bigO(\polylog K)$ amortized time \cite{holm2001poly}.
For laminar matroids
(which include uniform and partition matroids as special cases),
the worst-case time complexity for exact dynamic algorithms is bounded by
$\bigO(\log K)$ \cite{henzinger2023faster}.
For transversal matroids,
a $\bigO(K^{1.407})$-time exact dynamic algorithm is known \cite{vandenBrand2019dynamic}, while
a $(1+\err)$-approximation dynamic algorithm runs in $\bigO(\err^{-2}\sqrt{K})$ time \cite{gupta2013fully}.
We can safely assume that, after updating multiple arm weights, $\mathcal{A}$ returns an (approximate) maximum-weight basis in $\bigO(D)$ time, where $D$ is the rank of a matroid.



%% file: sections/4.subroutine.tex
\section{Dynamic Maintenance of Maximum-weight Base over Inner Product Weight}
\label{sec:dynamic}

In this section, we develop a sublinear-time sampling rule, which is used as a subroutine in {\tt FasterCUCB}.
Recall that
\emph{any} static algorithm that solves the linear maximization of Eq.~\eqref{eq:combucb} from scratch
requires at least $\Omega(K)$ time, which is computationally expensive.
To circumvent this issue,
we present a dynamic algorithm for maintaining an (approximate) maximum-weight base of a matroid
where arm weights change over time.
The next subsection begins with formalizing the problem setting.

\subsection{Problem Setting and Technical Result}

Consider the following problem setting:
Let $\matcls = ([K], \icls)$ be a matroid of rank $D$ over $K$ arms, and 
$\decls$ be the set of its bases.
Each arm $k \in [K]$ has a (nonnegative) two-dimensional vector
$\bd{f}_k = (\alpha_k, \beta_k) \in \reals_+^2$ referred to as a \emph{feature},
which may change as time goes by.
Given a (nonnegative) two-dimensional vector $\bd{q} \in \reals_+^2$ as a \emph{query},
we are required to find any (approximate) maximum-weight base of $\matcls$, where
arm $k$'s weight is given by projecting its feature onto $\bd{q}$, i.e.,
$\langle \bd{f}_k, \bd{q} \rangle$.

Observe that in the matroid semi-bandit setting,
each arm $k$'s feature $\bd{f}_k = (\alpha_k, \beta_k)$ corresponds to a pair of
the empirical reward estimate $\alpha_k = \hat{\mu}_k(t-1)$ and
radius $\beta_k = \frac{1}{\sqrt{N_k(t-1)}}$ of confidence interval, and
a query is $\bd{q} = (1, \lambda_t)$ at round $t$,
both of which change over rounds.

Hereafter, we make the following two assumptions.

\begin{assumption}
\label{asm:dynamic:bounds}
Lower and upper bounds,
denoted by $\alpha_\lb$ and $\alpha_\ub$ (resp. $\beta_\lb$ and $\beta_\ub$),
on the possible positive values of $\alpha_k$'s (resp.~$\beta_k$'s) at anytime are known;
namely, it always holds that
$\alpha_k \in \{0\} \cup [\alpha_\lb, \alpha_\ub]$ and $\beta_k \in \{0\} \cup [\beta_\lb, \beta_\ub]$.
The precise values of these bounds will be discussed in \cref{sec:regret-analysis}.
\end{assumption}

\begin{assumption}
\label{asm:dynamic:algorithm}
There exists a dynamic algorithm $\mathcal{A}$ for maintaining a $(1+\err)$-approximate maximum-weight base of $\matcls$
with arm weights changing over time, where
parameter $\err \in (0,1)$ specifies the approximation guarantee.
Denote by
$\mathcal{T}_{\mathrm{init}}(\mathcal{A}; \err)$ and
$\mathcal{T}_{\mathrm{update}}(\mathcal{A}; \err)$
the time complexity of $\mathcal{A}$ required for
initializing the data structure and
updating a single arm's weight,
respectively.
We can safely assume that after updating multiple arm weights,
$\mathcal{A}$ returns a maximum-weight base in $\bigO(D)$ time.
See \cref{sec:related-works} for existing implementations.
\end{assumption}

Our dynamic algorithm is parameterized by a \emph{precision parameter} $\epsilon \in (0,1)$, and
consists of the following three procedures:
\begin{description}
\item[\textbf{\textsc{Initialize}}:]
Given lower and upper bounds $[\alpha_\lb, \alpha_\ub]$ and $[\beta_\lb, \beta_\ub]$ as in \cref{asm:dynamic:bounds},
$K$ features $(\alpha_k, \beta_k)_{k \in [K]}$,
a matroid $\matcls = ([K], \icls)$,
a dynamic algorithm $\mathcal{A}$ for maximum-weight base maintenance as in \cref{asm:dynamic:algorithm}, and
a precision parameter $\epsilon$,
this procedure initializes the data structure used in the remaining two procedures.
\item[\textbf{\textsc{Find-Base}}:]
Given a query $\bd{q}$,
this procedure is supposed to return a $(1+\epsilon)$-approximate maximum-weight base of $\matcls$,
where arm $k$'s weight is defined as
$\langle \bd{f}_k, \bd{q} \rangle$ for the up-to-date $k$'s feature $\bd{f}_k$.
\item[\textbf{\textsc{Update-Feature}}:]
Given an arm $k$ and a new feature $\bd{f}'_k$,
this procedure reflects the change of arm $k$'s feature on the data structure.
\end{description}

\begin{remark}
Our problem setting is different
from a canonical setting of dynamic maximum-weight base maintenance in a sense that
the arm weights are revealed when a query is issued in \textbf{\textsc{Find-Base}}.
Consequently, existing dynamic algorithms may not be used directly.
\end{remark}

The technical result in this section is stated below.
\begin{theorem}[$*$]
\label{thm:dynamic}
There exist implementations of 
\textbf{\textsc{Initialize}},
\textbf{\textsc{Find-Base}}, and
\textbf{\textsc{Update-Feature}}
such that the following are satisfied:
\textbf{\textsc{Find-Base}} always returns
a $(1+\epsilon)$-approximate maximum-weight base of a matroid $\matcls$ with
arm $k$'s weight defined as $\langle \bd{f}_k, \bd{q} \rangle$
for an up-to-date $k$'s feature $\bd{f}_k$ and a query $\bd{q}$.
Moreover,
\textbf{\textsc{Initialize}} runs in
$\bigO(K+\poly{W} \cdot \mathcal{T}_{\mathrm{init}}(\mathcal{A}; \frac{\epsilon}{3}))$
time,
\textbf{\textsc{Find-Base}} runs in
$\bigO(\poly{W} + D)$
time, and
\textbf{\textsc{Update-Feature}} runs in
$\bigO(\poly{W} \cdot \mathcal{T}_{\mathrm{update}}(\mathcal{A}; \frac{\epsilon}{3}))$
time,
where
\begin{align}
    W = \bigO\left(
        \epsilon^{-1} \cdot \log\left(\frac{\alpha_\ub}{\alpha_\lb} \cdot \frac{\beta_\ub}{\beta_\lb}\right)
    \right).
\end{align}
\end{theorem}

\begin{remark}
The proof of \cref{thm:dynamic} can be easily adapted to the case when
(the update procedure of) dynamic algorithm $\mathcal{A}$ has only amortized complexity.
In such case, \cref{thm:dynamic} holds in the \emph{amortized} sense rather than the \emph{worst-case} sense.
\end{remark}

The remainder of this section is organized as follows:
In \cref{subsec:dynamic:rounding},
we apply a rounding technique to arm features to reduce the number of distinct features to consider,
in \cref{subsec:dynamic:multiple},
we investigate the representability of permutations induced by inner product weights to
deal with multiple queries efficiently, and
\cref{subsec:dynamic:together} finally develops
our dynamic algorithm for maximum-weight base maintenance.
All proofs of the lemmas appearing in this section are deferred to \cref{app:dynamic}.

\subsection{Rounding Arm Features}
\label{subsec:dynamic:rounding}
Here, we apply a rounding technique to arm features
so as to reduce the number of distinct features to consider.
Hereafter, let $\err \triangleq \frac{\epsilon}{3}$,
so that $(1+\err)^2 \leq 1+\epsilon$ for $\epsilon \in (0,1)$.
Define
\begin{align}
\label{eq:dynamic:R}
\begin{aligned}
W & \triangleq \max\left\{
    \left\lceil \log_{1+\err}\left(\frac{\alpha_\ub} {\alpha_\lb}\right) \right\rceil,
    \left\lceil \log_{1+\err}\left(\frac{\beta_\ub}{\beta_\lb}\right) \right\rceil
\right\}, \\
\mathbb{W} & \triangleq \{-\infty\} \cup [W]
= \{-\infty, 1, 2, 3, \ldots, W\}.
\end{aligned}
\end{align}
Since any features are within
$(\{0\}\cup[\alpha_\lb, \alpha_\ub]) \times (\{0\}\cup[\beta_\lb, \beta_\ub])$
as guaranteed by \cref{asm:dynamic:algorithm},
we shall partition the possible region of the features into $|\mathbb{W}|^2$ bins.
For each $q,r \in \mathbb{W}$,
define $\bin{q,r} \subset \reals_+^2$ as
\begin{align}
\label{eq:dynamic:bin}
\begin{aligned}
    \bin{q,r} \triangleq & (\alpha_\lb (1+\err)^{q-1}, \alpha_\lb(1+\err)^q] \\
    \times & (\beta_\lb (1+\err)^{r-1}, \beta_\lb (1+\err)^r],
\end{aligned} \\
\text{where }
(\alpha_\lb (1+\err)^{-\infty}, \alpha_\lb(1+\err)^{-\infty}] & \triangleq \{0\}, \\
(\beta_\lb (1+\err)^{-\infty}, \beta_\lb (1+\err)^{-\infty}] & \triangleq \{0\}.
\end{align}
Note that these bins are pairwise disjoint, and that $\cup_{q,r \in \mathbb{W}} \bin{q,r}$ covers
$(\{0\} \cup [\alpha_\lb, \alpha_\ub]) \times (\{0\} \cup [\beta_\lb, \beta_\ub])$; i.e.,
any possible feature belongs to a unique $\bin{q,r}$.
For each $q,r \in \mathbb{W}$,
let $\dom_{q,r} \in \reals_+^2$ denote the unique \emph{dominating point} of $\bin{q,r}$; namely,
\begin{align}
\label{eq:dynamic:dom}
    \dom_{q,r} \triangleq (\alpha_\lb(1+\err)^q, \beta_\lb(1+\err)^r).
\end{align}
For any feature $\bd{f}_k = (\alpha_k, \beta_k)$,
we will use $\dom(\bd{f}_k) = \dom(\alpha_k,\beta_k)$ to denote the dominating point $\dom_{q,r}$ such that
$\bd{f}_k \in \bin{q,r}$.
See \cref{fig:rounding} in \cref{app:dynamic} for illustration of
$\bin{q,r}$'s, $\dom_{q,r}$'s, and $\dom(\bd{f}_k)$'s.

Observe easily that for any feature $\bd{f}_k \in \reals_+^2$ and query $\bd{q} \in \reals^2_+$,
\begin{align}
\label{eq:dynamic:approx}
    \frac{1}{1+\err} \cdot \langle \dom(\bd{f}_k), \bd{q} \rangle
    < \langle \bd{f}_k, \bd{q} \rangle
    \leq \langle \dom(\bd{f}_k), \bd{q} \rangle.
\end{align}
By Eq.~\eqref{eq:dynamic:approx},
we can replace each arm's feature by its dominating point
without much deteriorating the quality of the (approximate) maximum-weight base,
as shown below.

\begin{lemma}[$*$]
\label{lem:dynamic:approx}
Let $\bd{f}_1, \ldots, \bd{f}_K
\in (\{0\}\cup[\alpha_\lb, \alpha_\ub]) \times (\{0\}\cup[\beta_\lb, \beta_\ub])$
be $K$ features,
$\bd{q} \in \reals^2_+$ be a query, and
$\bd{x}_\dom^*$ be a $(1+\err)$-approximate maximum-weight base of matroid $\matcls$
with arm $k$'s weight defined as $\langle \dom(\bd{f}_k), \bd{q} \rangle$.
Then, for any base $\bd{x}$ of $\matcls$, it holds that
\begin{align}
\sum_{k \in \supp{\bd{x}_\dom^*}} \langle \bd{f}_k, \bd{q} \rangle
\geq
\frac{1}{1+\epsilon} \cdot \sum_{k \in \supp{\bd{x}}} \langle \bd{f}_k, \bd{q} \rangle.
\end{align}
In particular, $\bd{x}_\dom^*$ is a $(1+\epsilon)$-approximate maximum-weight base
with arm $k$'s weight defined as $\langle \bd{f}_k, \bd{q} \rangle$.
\end{lemma}

\subsection{Handling Multiple Queries}
\label{subsec:dynamic:multiple}

\paragraph{From weighting to permutation.}
Now we deal with multiple queries.
Our idea is that, if two queries $\bd{q}_1, \bd{q}_2 \in \reals^2_+$
are ``very close'' to each other, 
then
they should derive the common maximum-weight base
(provided that features are fixed).
This intuition can be justified with respect to \emph{orderings} of arms.
For two total orders $\preceq$ and $\preceq^\circ$ over $[K]$,
we say that $\preceq^\circ$ is \emph{consistent with} $\preceq$ if 
$k \preceq^\circ k'$ implies $k \preceq k'$ for any $k \neq k'$.
The following fact is easy to confirm:

\begin{lemma}[$*$]
\label{lem:dynamic:unique-base}
Let $\bd{w} = (w_1, \ldots, w_K) \in \reals_+^K$ be $K$ arm weights and 
$\preceq$ be a total order over $[K]$ such that
$k \succeq k'$ if and only if $w_k \geq w_{k'}$. 
Let $\preceq^\circ$ be a total order over $[K]$ that is consistent with $\preceq$.
If $\bd{x}^\circ$ is a base of matroid $\matcls$ obtained by running the greedy algorithm
over any ordering of $[K]$ consistent with $\preceq^\circ$,
it is a maximum-weight base of $\matcls$ over arm $k$'s weight $w_k$; namely,
for any base $\bd{x}$ of $\matcls$,
\begin{align}
\langle \bd{x}^\circ, \bd{w} \rangle \geq \langle \bd{x}, \bd{w} \rangle.
\end{align}
\end{lemma}

\cref{lem:dynamic:unique-base} implies that 
any maximum-weight base can be obtained by running the greedy algorithm over some total order $\preceq^\circ$;
moreover, we can safely assume that $\preceq^\circ$ is \emph{strict}
(i.e., $k \prec^\circ k'$ or $k \succ^\circ k'$ for all $k \neq k'$), or equivalently,
a \emph{permutation} over $[K]$.
Our strategy for dealing with multiple queries is:
(1) we enumerate all possible permutations in advance, and
(2) we guess a permutation consistent with the arm weights determined based on a query.
To this end, the following question arises:
\emph{What kind of and how many permutations are representable given a fixed set of features?}

\paragraph{Characterizing representable permutations.}
To answer the above question, we characterize representable permutations.
Hereafter, let $\mathfrak{S}_K$ denote the set of all permutations over $[K]$, and
$\bd{f}_1, \ldots, \bd{f}_K \in \reals^2_+$ be any fixed, distinct $K$ features.
We say that
a query $\bd{q} \in \reals^2$ over $\bd{f}_1, \ldots, \bd{f}_K$ \emph{represents} a permutation $\pi \in \mathfrak{S}_K$ if
$\langle \bd{f}_{\pi(i)}, \bd{q} \rangle > \langle \bd{f}_{\pi(j)}, \bd{q} \rangle$ for all
$1 \leq i < j \leq K$,\footnote{
This definition does not allow ``ties''; i.e., no pair $k \neq k'$ satisfies
$\langle \bd{f}_k, \bd{q} \rangle = \langle \bd{f}_{k'}, \bd{q} \rangle$.
} and that
$\pi$ is \emph{representable} if such $\bd{q}$ exists.

For a permutation $\pi \in \mathfrak{S}_K$ to be representable,
we wish for some query $\bd{q} \in \reals^2$ to ensure that
for any $i < j$,
arm $\pi(i)$'s weight is (strictly) higher than arm $\pi(j)$'s weight.
This requirement is equivalent to
$\langle \bd{f}_{\pi(i)} - \bd{f}_{\pi(j)}, \bd{q} \rangle > 0$;
thus, if the following system of linear inequalities is feasible,
any of its solutions $\bd{q}$ represents $\pi$:
\begin{align}
\label{eq:dynamic:linear-system}
    \langle \bd{f}_{\pi(i)} - \bd{f}_{\pi(j)}, \bd{q} \rangle > 0
    \text{\;   for all \;} 1 \leq i < j \leq K.
\end{align}
Observe now that the above system is feasible \emph{if and only if}
the intersection of $\mathcal{P}_{i,j}$ for all $i < j$ is nonempty, where
$\mathcal{P}_{i,j}$ is an open half-plane defined as
\begin{align}
    \mathcal{P}_{i,j}
    \triangleq \{ \bd{q} \in \reals^2 : \langle \bd{f}_{\pi(i)} - \bd{f}_{\pi(j)}, \bd{q} \rangle > 0 \}.
\end{align}
Because each $\mathcal{P}_{i,j}$ is obtained by dividing $\reals^2$ by
a unique line that
is orthogonal to line $\overleftrightarrow{\bd{f}_{\pi(i)} \bd{f}_{\pi(j)}}$ and
intersects the origin $\bd{0}$,
the set of feasible solutions for Eq.~\eqref{eq:dynamic:linear-system}
is equal to
(the interior of) a \emph{polyhedral cone} defined by the boundaries of a particular pair of $\mathcal{P}_{i,j}$'s.

Here, we characterize the representable permutations by the concept of arrangement of lines.
Let
$\mathcal{L} \triangleq \{\bd{l}_1, \ldots, \bd{l}_{K \choose 2} \}$ denote ${K \choose 2}$ lines,
each of which is orthogonal to line $\overleftrightarrow{\bd{f}_k\bd{f}_{k'}}$
for some $k \neq k'$ and intersects $\bd{0}$.
Given such $\mathcal{L}$,
a \emph{cell $\mathcal{C}$ in arrangement of $\mathcal{L}$} is 
defined as a maximum connected region of $\reals^2$ that does not intersect with $\mathcal{L}$
(which is the interior of a polyhedral cone).
Then, for each cell $\mathcal{C}$,
every query $\bd{q}$ in $\mathcal{C}$ represents the same permutation $\pi_{\mathcal{C}} \in \mathfrak{S}_K$
depending only on $\mathcal{C}$; namely,
\emph{there is a bijection between
the representable permutations and the cells in arrangement of $\mathcal{L}$}.
See \cref{fig:representable} in \cref{app:dynamic} for illustration.

With this connection in mind,
we demonstrate that reserving a single vector for each cell suffices to cover all representable permutations.
A \emph{minimum hitting set} of the cells in arrangement of $\mathcal{L}$
is defined as
any minimum set $\mathcal{H}$ of
vectors in $\reals^2$ such that
$\mathcal{H}$ and each cell have a non-empty intersection.

\begin{lemma}[$*$]
\label{lem:dynamic:covering}
Let $\mathcal{H}$
be a minimum hitting set of the cells in arrangement of $\mathcal{L}$.
Then, for any query $\bd{q} \in \reals^2$,
there exists a vector $\bd{h} \in \mathcal{H}$ such that
for any $k \neq k'$,
\begin{align}
\langle \bd{f}_k, \bd{h} \rangle > \langle \bd{f}_{k'}, \bd{h} \rangle
\implies
\langle \bd{f}_k, \bd{q} \rangle \geq \langle \bd{f}_{k'}, \bd{q} \rangle.
\end{align}
\end{lemma}

\cref{lem:dynamic:covering} along with \cref{lem:dynamic:unique-base}
ensure that
for any query $\bd{q} \in \reals^2$,
there is a vector $\bd{h}$ in $\mathcal{H}$ such that
a maximum-weight base with arm $k$'s weight $\langle \bd{f}_k, \bd{h} \rangle$ is
a maximum-weight base with arm $k$'s weight $\langle \bd{f}_k, \bd{q} \rangle$.

\paragraph{Generating a minimum hitting set.}
Subsequently, we generate a minimum hitting set.
One may think that it requires exponentially long time because the number of permutations in $\mathfrak{S}_K$ is $K!$.
However,
it turns out that the number of cells in arrangement of $\mathcal{L}$ is $\bigO(K^2)$
(i.e., so is the number of representable permutations), and
a minimum hitting set can be constructed in $\poly{K}$ time.

\begin{lemma}[$*$]
\label{lem:dynamic:hyperplane-arrangement}
The number of cells in arrangement of $\mathcal{L}$ is at most $\bigO(K^2)$.
Moreover, we can generate a minimum hitting set in $\poly{K}$ time
(by using \cref{alg:dynamic:hitting-set} in \cref{app:dynamic}).
\end{lemma}

\subsection{Putting It All Together: Algorithm Description and Complexity}
\label{subsec:dynamic:together}
We are now ready to implement the three procedures.
We here stress that
applying either of the feature rounding or minimum hitting set technique \emph{separately}
does not make sense:
On one hand, if we only apply feature rounding,
we would have to recompute each arm's weight \emph{every time} a query is issued,
which is expensive.
On the other hand, if the minimum hitting set technique is only applied (to raw features $\bd{f}_k$'s),
then
a minimum hitting set $\mathcal{H}$ cannot be constructed in advance due to a dynamic nature of features, and
its size would be $\bigO(K^2)$, which is prohibitive.

By applying both techniques,
(1) we know \emph{a priori} the set of possible dominating points, whose size $\bigO(W^2)$ depends \emph{only} on $W$;
moreover, 
(2) we can create a minimum hitting set $\mathcal{H}$ of size $\bigO(W^4)$ beforehand at initialization.
Pseudocodes of \textbf{\textsc{Initialize}}, \textbf{\textsc{Find-Base}}, and \textbf{\textsc{Update-Feature}}
are described in \cref{alg:dynamic:init,alg:dynamic:base,alg:dynamic:update}, respectively.
The proof of \cref{thm:dynamic}
follows from 
\cref{lem:dynamic:approx,lem:dynamic:unique-base,lem:dynamic:covering,lem:dynamic:hyperplane-arrangement},
whose details are deferred to \cref{app:dynamic}.

In \textbf{\textsc{Initialize}},
we construct a hitting set $\mathcal{H}$ of $\dom_{q,r}$'s and $\frac{1}{1+\err}\cdot \dom_{q,r}$'s
rather than solely of $\dom_{q,r}$'s,
which incurs a constant-factor blowup in the time complexity.
Though this change is not needed in the proof of \cref{thm:dynamic},
the following immediate corollary of \cref{lem:dynamic:covering}
is crucial in the regret analysis of \cref{sec:regret-analysis}.
\begin{corollary}[$*$]
\label{cor:dynamic:covering-new}
    Let $\mathcal{H}$ be a minimum hitting set constructed in \cref{alg:dynamic:init}.
    Then, for any query $\bd{q} \in \reals^2$,
    there exists a vector $\bd{h} \in \mathcal{H}$ such that for any
    $\dom = \dom_{q,r}$ and $\dom' = \dom_{q',r'}$ with $q,r,q',r' \in \mathbb{W}$,
    \begin{align*}
        \langle \dom, \bd{h} \rangle > \langle \dom', \bd{h} \rangle
        \implies
        \langle \dom, \bd{q} \rangle \geq \langle \dom', \bd{q} \rangle, \\
        \langle \dom, \bd{h} \rangle > \frac{\langle \dom', \bd{h} \rangle}{1+\eta}
        \implies
        \langle \dom, \bd{q} \rangle \geq \frac{\langle \dom', \bd{q} \rangle}{1+\eta}.
    \end{align*}
\end{corollary}

\begin{algorithm}[h!]
\caption{\textbf{\textsc{Initialize}}.}
\label{alg:dynamic:init}
\footnotesize
    \KwIn{lower and upper bounds $[\alpha_\lb, \alpha_\ub]$ and $[\beta_\lb, \beta_\ub]$;
    $K$ features $(\bd{f}_k)_{k \in [K]}$;
    precision parameter $\epsilon \in (0,1)$.
    }
    Define $\mathbb{W}$ by Eq.~\eqref{eq:dynamic:R};
    
    \For{each $q,r \in \mathbb{W}$}{
        Define $\bin{q,r}$ and $\dom_{q,r}$ by Eqs.~\eqref{eq:dynamic:bin} and \eqref{eq:dynamic:dom};
    }
    Define $\eta \triangleq \frac{\epsilon}{3}$;
    
    Construct a minimum hitting set $\mathcal{H}$
    of size $\bigO(W^4)$ for $\dom_{q,r}$'s
    and $\frac{1}{1+\eta} \cdot \dom_{q,r}$'s
    by \cref{lem:dynamic:hyperplane-arrangement};
    
    \For{each $\bd{h} \in \mathcal{H}$}{
        Create an instance $\mathcal{A}_{\bd{h}}$ of dynamic maximum-weight base algorithm 
        with
        precision parameter $\eta \triangleq \frac{\epsilon}{3}$,
        $\matcls$, and
        arm $k$'s weight $\langle \dom(\bd{f}_k), \bd{h} \rangle$;
    }
\end{algorithm}

\begin{algorithm}[h!]
\caption{\textbf{\textsc{Find-Base}}.}
\label{alg:dynamic:base}
\footnotesize
    \KwIn{query $\bd{q} \in \reals_+^2$.}
    Find $\bd{h} \in \mathcal{H}$
    s.t.~$\bd{q}$ and $\bd{h}$ belong to (the closure of) the same cell in arrangement of $\mathcal{V}$;
    
    Call $\mathcal{A}_{\bd{h}}$ and
    return the maximum-weight base $\bd{x}^\circ$ of $\matcls$
    with arm $k$'s weight $\langle \dom(\bd{f}_k), \bd{h} \rangle$;
\end{algorithm}

\begin{algorithm}[h!]
\caption{\textbf{\textsc{Update-Feature}}.}
\label{alg:dynamic:update}
\footnotesize
    \KwIn{arm $k \in [K]$; new feature $\bd{f}'_k \in \reals_+^2$.}
    \For{each $\bd{h} \in \mathcal{H}$}{
        Change arm $k$'s weight stored in $\mathcal{A}_{\bd{h}}$
        to $\langle \dom(\bd{f}'_k), \bd{h} \rangle$;
    }
\end{algorithm}

%% file: sections/5.regret.tex
\section{Our Proposed Algorithm: {\tt FasterCUCB}}\label{sec:regret-analysis}
In this section, we present {\tt FasterCUCB} in Algorithm~\ref{alg:faster-cucb}, which uses procedures introduced in Section~\ref{sec:dynamic}.

The purpose of initialization procedure is to ensure each arm is pulled at least once. It takes at most $K$ rounds, and in each round, the computation of $\istar{\be_k}$ takes $\bigO(K\cdot\mathcal{T}_{\mathrm{member}})$ time because the permutation $\gamma$ in Algorithm~\ref{alg:greedy} can be specified explicitly as $\gamma(j)=k$ if $j=1$, $\gamma(j)=j-1$ if $2\leq j<k$, and $\gamma(j)=j+1$ if $j\geq k$. So, it only require to compute at most $K$ membership tests. 
After the initialization, the computation of each round $t$ consists of one call to \textbf{\textsc{Find-base}} for computing the action $\bx(t)$, the update of $\hat{\mu}_k(t)$ and $N_k(t)$ for each $k\in\supp{\bx(t)}$, and one call to \textbf{\textsc{Update-Feature}} for updating the feature of each arm $k\in\supp{\bx(t)}$ stored in the instances of the dynamic maxium-weight base algorithm.

\begin{algorithm}[h!]
\caption{\texttt{FasterCUCB}}
\label{alg:faster-cucb}
\footnotesize
    \KwIn{the total number of rounds $T$, $\lambda_t$, and $m\in\ints$}
    \SetKwProg{Init}{Initialization}{:}{}
    \Init{}{
        $t=0$;
        
        \While{$\exists k\in[K]$ such that $N_k(t)=0$}
        {
            Pull $\istar{\be_k}$;
            $t = t+1$;
        }
        
        \textbf{\textsc{Initialize}}$\left(a, b, \frac{1}{\sqrt{T}}, 1, \left(\hat{\mu}_k(t), N_k(t)\right)_{k\in[K]}, \frac{1}{\log^m T}\right)$
    }
    \While{$t < T$}{
        $\bx(t)\leftarrow$\textbf{\textsc{Find-Base}}$\left((1,\lambda_t)\right)$;

        Pull $\bx(t)$ and receive $y_k(t)\sim\nu_k$ for each $k\in\supp{\bx(t)}$;
        
        \For{$k\in\supp{\bx(t)}$}{
            $N_k(t)\leftarrow N_k(t-1)+1$;
            
            $\hat{\mu}_k(t)\leftarrow \frac{t-1}{t}\hat{\mu}_k(t-1)+\frac{1}{t}y_k(t)$;
            
            \textbf{\textsc{Update-Feature}}$\left(k, \left(\mu_k(t),\frac{1}{\sqrt{N_k(t)}}\right)\right)$;
        }
        $t = t+1$;
    }    
\end{algorithm}

\subsection{Per-round Time Complexity}
By Theorem~\ref{thm:dynamic}, one call to \textbf{\textsc{Find-Base}} takes 
$\bigO\left(\poly{W}+D\right)$ and $D$ calls to \textbf{\textsc{Update-Feature}} take $\bigO(D\,\poly{W}\,\mathcal{T}_{\mathrm{update}}(\mathcal{A}; \frac{\epsilon}{3}))$.
Since \[
W
=\bigO\left(\log^mT\log\left(\frac{b}{a}\sqrt{T}\right)\right)
=\bigO\left(\log^{m+1}T\right),
\]
the per-round time complexity of Algorithm~\ref{alg:faster-cucb} is
\[\bigO\left(
    D\,\polylog{T}\,\mathcal{T}_{\mathrm{update}}\left(\mathcal{A}; \frac{\epsilon}{3}\right)
\right)\].
Here, we will set $\epsilon = \frac{1}{\log^m T}$ for the regret analysis.


\subsection{Regret Upper Bound}
\paragraph{Notation.}
Fix $\bmu\in\Lambda$ and $\ist\in\argmax_{\bx\in\decls}\inner{\bmu}{\bx}$. We introduce a few notation.
Let $\{\overline{j}\}_{j=1}^D$ be the permutation of $\supp{\ist}$ such that $\mu_{\overline{1}}\geq\cdots\geq\mu_{\overline{D}}$.
Define $\gap{j,k}\triangleq\mu_j-\mu_k$ and $d_k\triangleq\max\{j\in[D]:\gap{\overline{j},k}>0\}$ for $j\in\supp{\ist}$ and $k\notin\supp{\ist}$, 
and $\gap{\min}\triangleq\min_{k\notin\supp{\ist}}\gap{\overline{d_k},k}$.

\begin{theorem}\label{thm:regret-noanytime}
Let $\lambda_t=\sqrt{1.5(b-a)^2\log t}$ and $m\in\ints$.
Define $T_0\triangleq\max\{K,\exp((\frac{b}{\gap{\min}})^{\frac{1}{m}})\}$.
For $T\in\ints$, the expected regret of Algorithm~\ref{alg:faster-cucb} is upper bounded by 
\begin{align*}
R(T) 
& \leq \sum_{k\notin\supp{\ist}}\left(\sum_{j=1}^{d_k}\gap{\overline{j},k}T_0
+\frac{12\gap{\overline{d_k},k}(b-a)^2\log T}{\left(\frac{\mu_{\overline{d_k}}}{1+\log^{-m}T}-\mu_k\right)^2}\right)\\
& +\sum_{k\notin\supp{\ist}}\sum_{j=1}^{d_k}\gap{\overline{j},k}\left(\frac{1}{T}+\frac{\pi^2}{6}\right) +DbT_0 .
\end{align*}
\end{theorem}

As a consequence of Theorem~\ref{thm:regret-noanytime}, setting $T\to\infty$ yields: 
\[
\lim_{T\to\infty}\frac{R(T)}{\log T}
\leq\sum_{k\notin\supp{\ist}}\frac{12(b-a)^2}{\gap{\overline{d_k},k}}
\leq\bigO\left(\frac{K-D}{\gap{\min}}\right),
\]
which matches Theorem 4 in \cite{kveton2014matroid}, $\liminf_{T\to\infty}\frac{R(T)}{\log T}=\Omega(\frac{K-D}{\gap{\min}})$, asymptotically up to a constant factor.
Note that {\tt FasterCUCB} is faster than {\tt CUCB} when $\gap{\min}=\Omega(\frac{1}{\polylog{K}})$ and when $T=\poly{K}$. Also, similar to \cite{cuvelier2021statistically}, our per-round time complexity also goes to infinity as $T\to\infty$, one way to address this issue is to use {\tt CUCB} when the per-round time complexity of ours is larger than that of {\tt CUCB}.


\paragraph{Useful lemmas.}
Here we present two lemmas that will be used to show Theorem~\ref{thm:regret-noanytime} in Section~\ref{sec:proof-regret-noanytime}.
First, inspired by \citet{kveton2014matroid}, we define a bijection $g_t$ from $\supp{\ist}$ to $\supp{\bx(t)}$ with the following properties:
\begin{lemma}\label{lem:gt-property}
There exists a bijection $g_t:\supp{\ist}\to\supp{\bx(t)}$ such that (i) $g_t(j)=j$ for $j\in\supp{\ist}\cap\supp{\bx(t)}$; (ii) for any $j\in\supp{\ist}\backslash\supp{\bx(t)}$, 
\[
x_{g_t(j)}(t)=1
\implies
\inner{\dom(\bdf_{g_t(j)})}{\bh}\geq\frac{\inner{\dom(\bdf_j)}{\bh}}{1+\frac{1}{3\log^mT}}.
\]
\end{lemma}
The proof of Lemma~\ref{lem:gt-property} is in Appendix~\ref{proof:gt-property}, where an explicit construction of $g_t$ is provided. 
Property (i) allows us to decompose the instantaneous regret 
$\inner{\ist-\bx(t)}{\bmu}=\sum_{k\notin\supp{\ist}}\sum_{j\in\supp{\ist}}\gap{j,k}\ind{g_t(j)=k}$, and Property (ii), Lemma~\ref{lem:gt-property}, allows us to derive a bound of $\sum_{t=1}^T\ind{g_t(j)=k}$ that relates with UCB values. 

Second, for technical reasons, we need the precision parameter $\epsilon=\log^{-m}T$ to be small enough so that $\gap{i,j}$ and $\mu_i-(1+\epsilon)\mu_j$ have the same sign. The below lemma (proved in Appendix~\ref{proof:regret-T0}) gives the threshold to make it happen: 
\begin{lemma}\label{lem:regret-T0}
Let $\epsilon<\frac{\gap{\min}}{b}$. Then, for any $i\in\supp{\ist}$ and any $j\notin\supp{\ist}$, 
$\mu_i-\mu_j>0 \implies \frac{\mu_i}{1+\epsilon}-\mu_j>0$.
\end{lemma}

\subsection{Proof of Theorem~\ref{thm:regret-noanytime}}\label{sec:proof-regret-noanytime}
For $T\leq T_0$, $R(T)$ is trivially bounded by $T\inner{\bmu}{\ist}\leq DbT_0$.
In the following, we assume $T>T_0$.

As $g_t$ is a bijection from $\supp{\ist}$ to $\supp{\bx(t)}$ and $g_t(j)=j$ for $j\in\supp{\ist}\cap\supp{\bx(t)}$, we can rewrite 
\begin{align*}
\expectation{}{\inner{\ist-\bx(t)}{\bmu}} &=\sum_{k\notin\supp{\ist}}\sum_{j\in\supp{\ist}}\gap{j,k}\expectation{}{\ind{g_t(j)=k}}\\
& \leq \sum_{k\notin\supp{\ist}}\sum_{j=1}^{d_k}\gap{\overline{j},k}\expectation{}{\ind{g_t(\overline{j})=k}}
\end{align*}
so that the expected regret is bounded from the above by: 
\begin{align*}
R(T)
& \leq \sum_{k\notin\supp{\ist}}\sum_{j=1}^{d_k}\gap{\overline{j},k}\expectation{}{\sum_{t=1}^T\ind{g_t(\overline{j})=k}}\\
& =\sum_{k\notin\supp{\ist}}\sum_{j=1}^{d_k}\gap{\overline{j},k}\left((I)_{\overline{j},k}+(II)_{\overline{j},k}\right),
\end{align*}
where 
$\begin{cases}
(I)_{\overline{j},k} =\sum_{t=1}^T\expectation{}{\ind{g_t(\overline{j})=k, N_k(t)\leq n_{\overline{j},k}}}\\
(II)_{\overline{j},k} =\sum_{t=1}^T\expectation{}{\ind{g_t(\overline{j})=k, N_k(t)> n_{\overline{j},k}}}
\end{cases}$
and $n_{j,k}=\max\left\{\frac{6(b-a)^2\log T}{(\frac{\mu_j}{1+\log^{-m}T}-\mu_k)^2},T_0\right\}$.
The proof is completed by bounding related terms of $(I)_{\overline{j},k}$ and $(II)_{\overline{j},k}$ by Lemma~\ref{lem:regret-anytime-part1} (proved in Appendix~\ref{proof:regret-anytime-part1})
and Lemma~\ref{lem:regret-part2}.
\begin{lemma}\label{lem:regret-anytime-part1}
Let $k\notin\supp{\ist}$ and $j\in[d_k]$. For $T>T_0$,
\[
\sum_{j=1}^{d_k}\gap{\overline{j},k}(I)_{\overline{j},k} \leq 
\sum_{j=1}^{d_k}\gap{\overline{j},k}T_0
+\frac{12(b-a)^2\gap{\overline{d_k},k}\log T}{(\frac{\mu_{\overline{d_k}}}{1+\log^{-m}T}-\mu_k)^2}.
\]
\end{lemma}
\begin{lemma}\label{lem:regret-part2}
Let $k\notin\supp{\ist}$ and $j\in[d_k]$. For $T>T_0$,
\[
(II)_{\overline{j},k} 
\leq \frac{1}{T}+\frac{\pi^2}{6}.
\]
\end{lemma}
\paragraph{Proof sketch:}
See Appendix~\ref{proof:regret-part2} for the entire proof.
Let $\epsilon\triangleq\frac{1}{\log^mT}$.
First, we claim: 
\begin{equation}\label{eq:regret-e0}
g_t(\overline{j})=k
\implies
u_{k}(N_{k}(t-1),T)
\geq \frac{\min_{s<t}u_{\overline{j}}(s,t)}{1+\epsilon},
\end{equation}
where $u_k(s,t)=\tilde{\mu}_k(s)+\frac{\lambda_t}{\sqrt{s}}$ and $\tilde{\mu}_k(t)=\frac{1}{t}\sum_{s=1}^ty_k(s)$.

\underline{Show Eq.~\eqref{eq:regret-e0}}: 
Observe that $g_t(\overline{j})=k$ implies:
\begin{align}
\left(1+\frac{\epsilon}{3}\right)\inner{\bdf_{k}}{\bq}
& \geq \inner{\dom(\bdf_{k})}{\bq}\nonumber\\
& \geq
\frac{\inner{\dom(\bdf_{\overline{j}})}{\bq}}{1+\frac{\epsilon}{3}}
\geq
\frac{\inner{\bdf_{\overline{j}}}{\bq}}{1+\frac{\epsilon}{3}},\label{eq:regret-e01}
\end{align}
where Eq.~\eqref{eq:dynamic:approx} is used in the first and the last inequality, and the second uses Lemma~\ref{lem:gt-property} and Corollary~\ref{cor:dynamic:covering-new}.
By $(1+\frac{\epsilon}{3})^2\leq 1+\epsilon$ and expanding $\bdf_k=(\hat{\mu}_k(t-1),\frac{1}{\sqrt{N_k(t-1)}})$ and $\bq=(1,\lambda_t)$, we derive from \eqref{eq:regret-e01} that: 
\[
u_{k}(N_{k}(t-1),t)\\
\geq \frac{u_{\overline{j}}(N_{\overline{j}}(t-1),t)}{1+\epsilon},
\]
and further by $\log T>\log t$ and $N_{\overline{j}}(t-1)\in[t-1]$, 
\begin{align*}
u_{k}(N_{k}(t-1),T) \geq \frac{u_{\overline{j}}(N_{\overline{j}}(t-1),t)}{1+\epsilon}
\geq \frac{\min_{s<t}u_{\overline{j}}(s,t)}{1+\epsilon},
\end{align*}
which shows Eq.~\eqref{eq:regret-e0}.
Second, define 
\[
\mathcal{T}_{\overline{j},k}=\{t\in\{n_{\overline{j},k}+1,\cdots,T\}:g_t(\overline{j})=k,N_k(t-1)>n_{\bar{j},k}\}.
\]
From Eq.~\eqref{eq:regret-e0}, $(II)_{\overline{j},k}$ is upper bounded by
\begin{align}
 & \expectation{}{\sum_{t\in\in\mathcal{T}_{\overline{j},k}}\ind{u_k(N_k(t-1),T) \geq \frac{\min_{s<t}\{u_{\overline{j}}(s,t)\}}{1+\epsilon}}}\nonumber\\
& \leq
\expectation{}{\sum_{t\in\mathcal{T}_{\overline{j},k}}\sum_{s<t}
\left(
    \ind{\mathcal{A}_{1,t,s}}
    +\ind{\mathcal{A}_{2,t,s}}
    +\ind{\mathcal{A}_{3,t,s}}
\right)},\label{eq:regret-part2-sketch-e1}
\end{align}
where 
$\begin{cases}
\mathcal{A}_{1,t,s}=\left\{\tilde{\mu}_k(N_k(t-1)) \geq \mu_k+\frac{\lambda_T}{\sqrt{N_k(t-1)}}\right\}\\
\mathcal{A}_{2,t,s}=\left\{\mu_{\overline{j}} \geq \tilde{\mu}_{\overline{j}}(s)+\frac{\lambda_t}{\sqrt{s}}\right\}\\
\mathcal{A}_{3,t,s}=\left\{\mu_k+\frac{2\lambda_T}{\sqrt{N_k(t-1)}} > \frac{\mu_{\overline{j}}}{1+\epsilon}\right\}
\end{cases}
$.
See Appendix~\ref{proof:regret-part2} for the derivation of Eq.~\eqref{eq:regret-part2-sketch-e1}.
Observe when $t\in\mathcal{T}_{\overline{j},k}$,
\[
\ind{\mathcal{A}_{3,t,s}}\leq \ind{\mu_k+\frac{2\lambda_T}{\sqrt{n_{\overline{j},k}+1}}>\frac{\mu_{\overline{j}}}{1+\epsilon}}=0,
\]
where the inequality is because $N_k(t-1)>n_{\overline{j},k}$, 
and the equality is because  
\[
n_{\overline{j},k}
\geq 
\frac{4\lambda_T^2}{(\frac{\mu_{\overline{j}}}{1+\epsilon}-\mu_k)^2}
\implies
\frac{4\lambda_T^2}{n_{\overline{j},k}+1} < \left(\frac{\mu_{\overline{j}}}{1+\epsilon}-\mu_k\right)^2,
\]
and also $\frac{\mu_{\overline{j}}}{1+\epsilon}-\mu_k>0$ which is ensured by Lemma~\ref{lem:regret-T0} as $T>T_0$.
Finally, from Eq.~\eqref{eq:regret-part2-sketch-e1} and using Hoeffding's inequality, we get
\begin{align*}
(II)_{\overline{j},k} 
& \leq \expectation{}{\sum_{t\in\mathcal{T}_{\overline{j},k}}\sum_{s<t}\left(\ind{\mathcal{A}_{1,t,s}}
    +\ind{\mathcal{A}_{2,t,s}}\right)}\\
& \leq \sum_{t=n_{\overline{j},k}+1}^T\sum_{s<t}\left(e^{-3\log T}+e^{-3\log t}\right).
\end{align*}
See Appendix~\ref{proof:regret-part2} for the derivation of the second inequality.
The proof is completed by evaluating
\[
\begin{cases}
\sum_{t=1}^T\sum_{s<t}e^{-3\log T}\leq\sum_{t=1}^T\frac{t}{T^3}
\leq \frac{T(T+1)}{2T^3}
\leq\frac{1}{T}\\
\sum_{t=1}^T\sum_{s<t}e^{-3\log t}
\leq \sum_{t=1}^{\infty}\frac{t}{t^3} 
\leq \sum_{t=1}^{\infty}\frac{1}{t^2} \leq \frac{\pi^2}{6}
\end{cases}.
\]
\hfill$\square$

%% file: sections/6.conclusion.tex
\section{Conclusion}\label{sec:conclusion}
In this paper, we have presented {\tt FasterCUCB}, the first sublinear-time algorithm for matroid semi-bandits. Several possible future directions. First, one might extend our approach to speed up UCB-style algorithms for different problems such as combinatorial best-arm identification \cite{chen2014combinatorial,du2021combinatorial} and nonstationary semi-bandits \cite{zhou2020near,chen2021combinatorial}. Second, another direction is to study the possibility of speeding up other forms of weights, such as those derived from gradients \cite{tzeng2023closing} and those in the follow-the-perturbed-leader algorithm \cite{neu2016importance}.

\section*{Acknowledgement}
Ruo-Chun Tzeng’s research is supported by the ERC Advanced
Grant REBOUND (834862).
Kaito Ariu's research is supported by JSPS KAKENHI Grant No. 23K19986.

\section*{Impact Statement}
This paper develops an algorithm for matroid semi-bandits. 
The societal consequences of this work indirectly come from the applications of matroid semi-bandits, and none of which we think should be discussed here.

%% file: app_notation.tex
\section{Notation}\label{app:notation}
\kaitocomment{I think right now they are not limited to Section~\ref{sec:pre} Section~\ref{sec:dynamic} Section~\ref{sec:regret-analysis}?\\RC: I'll remove the sentence}
\begin{table}[h!]
    \centering
    \begin{tabular}{l|l}
    \toprule
    \multicolumn{2}{l}{\bf Problem setting}\\
    \toprule
        $K$ & the number of arms\\
        $\decls$ & the bases of the given matroid $([K],\icls)$\\
        $D$ & $\max_{\bx\in\decls}\norm{\bx}_0$\\
        $\bmu$ & the mean vector of the $K$ arms $\nu_1,\cdots,\nu_K$\\
        $\istar{\bmu}$ & an action attaining $\max_{\bx\in\decls}\inner{\bmu}{\bx}$\\
    \toprule
    \multicolumn{2}{l}{\bf Notation related to {\tt FasterCUCB}}\\
    \toprule
        $N_k(t)$ & the number of arm pulls of arm $k$\\
        $\bx(t)$ & the action selected by the algorithm at round $t$\\
        $\by(t)$ & the reward vector at round $t$\\
        $\hat{\mu}_k(t)$ & the empirical reward $\frac{1}{N_k(t)}\sum_{s=1}^ty_k(t)\ind{x_k(t)=1}$ of arm $k$ at round $t$\\
        $\lambda_t$ & the parameter that controls the confidence interval\\
    \toprule
    \multicolumn{2}{l}{\bf Notation related to dynamic algorithm}\\
    \toprule
        $\bd{f}_k = (\alpha_k, \beta_k)$ & a nonnegative two-dimensional feature of arm $k$ \\
        $\bd{q}$ & a nonnegative two-dimensional query \\
        $(\alpha_\lb, \alpha_\ub)$ & lower and upper bounds of $\alpha_k$'s\\
        $(\beta_\lb, \beta_\ub)$ &
            lower and upper bounds of $\beta_k$'s \\
        $W$ & (the square root of) the number of bins \\
        $\bin{q,r}$ & bins that partition the possible region of the features \\
        $\dom_{q,r}$ & dominating point of $\bin{q,r}$ \\
        $\dom(\bd{f}_k)$ & dominating point of $\bin{q,r}$ to which $\bd{f}_k$ belongs \\
        $\mathcal{L} = \{\bd{l}_1, \ldots, \bd{l}_{K \choose 2}\}$ &
        the set of ${K \choose 2}$ lines, each of which is orthogonal to line $\overleftrightarrow{\bd{f}_k\bd{f}_{k'}}$ for some $k \neq k'$ and intersects $\bd{0}$ \\
        $\mathcal{H}$ & a minimum hitting set of the cells in arrangement of $\mathcal{L}$ \\
    \toprule
    \multicolumn{2}{l}{\bf Notation related to regret analysis}\\
    \toprule
        $\{\overline{j}\}_{j=1}^D$ & the permutation of $\supp{\ist}$ such that $\mu_{\overline{1}}\geq\cdots\geq\mu_{\overline{D}}$
        \\
        $\epsilon$ & the precision parameter which is set to $\frac{1}{\log^mT}$ in {\tt FasterCUCB} (Algorithm~\ref{alg:faster-cucb})\\
        $g_t(j)$ & the mapping from $\supp{\ist}$ to $\supp{\bx(t)}$ such that \\
        & $\quad$ (i) $g_t(j)=j$ if $j\in\supp{\ist}\cap\supp{\bx(t)}$\\
        &  $\quad$ (ii) $x_{g_t(j)}(t)=1$ implies $\inner{\dom(\bdf_{g_t(j)})}{\bh}\geq\frac{1}{1+\frac{\epsilon}{3}}\inner{\dom(\bdf_j)}{\bh}$\\
        $\gap{j,k}$ & the difference $\mu_j-\mu_k$ between arm $j$'s and arm $k$'s expected reward\\
        $\gap{\min}$ & the smallest positive gap $\gap{i,j}$ between any pair of $i\in\supp{\ist}$ and $j\notin\supp{\ist}$ \\
        $d_k$ & the largest $j\in[D]$ such that $\gap{\delta(j),k}>0$\\
        $\tilde{\mu}_k(t)$ & the average $\frac{1}{t}\sum_{s=1}^ty_k(s)$ of rewards of arm $k$ in the first $t$ rounds\\
        $u_k(s,t)$ & the UCB value of $\tilde{\mu}_k(s)+\frac{\lambda_t}{\sqrt{s}}$ under $s$ samples of arm $k$ and with confidence parameter $\lambda_t$\\
    \bottomrule
    \end{tabular}
    \caption{Table of notation.}
    \label{tab:regret_table}
\end{table}

%% file: app_related.tex
\section{Further Related Works}\label{sec:further-related-works}
In this section, we review relevant literatures on combinatorial semi-bandits and sublinear-time bandits. We focus on the stochastic setting. 
For ease of comparison, we assume the best action $\ist\in\argmax_{\bx\in\decls}\inner{\bx}{\bmu}$ is unique, and define  
$\gap{\min}\triangleq\min_{j,k\in[K]:i^{\star}_j=1, i^{\star}_k=0,\mu_j-\mu_k>0}(\mu_j-\mu_k)$, and $\gap{}\triangleq\min_{\bx\neq\ist:\inner{\ist-\bx}{\bmu}>0}\inner{\ist-\bx}{\bmu}$.

\paragraph{Matroid semi-bandits.}
\citet{kveton2014matroid} showed an instance such that any uniformly good algorithm\footnote{A uniformly good algorithm has the expected regret $R(T)=o(T^{\alpha})$ hold for any $\alpha>0$.} suffer $R(T)=\Omega\left(\frac{(K-D)\log T}{\gap{\min}}\right)$.
They also showed that {\tt CUCB} \cite{gai2012combinatorial,chen2013combinatorial} have a regret bound scaling as $\bigO\left(\frac{(K-D)\log T}{\gap{\min}}\right)$. 
\citet{talebi2016optimal} showed an instance-specific lower bound $\liminf_{T\to\infty}\frac{R(T)}{\log T}\geq c(\bmu)$ for uniformly good algorithms, where $c(\bmu)$ is the optimum of a semi-infinite linear program \cite{graves1997asymptotically,combes2015combinatorial}, and proposed {\tt KL-OSM} whose regret upper bound matches this lower bound.
The per-round compleixty of {\tt KL-OSM} is $K$ line search for generating the indices plus the time for solving a linear maximization.
Both {\tt CUCB} and {\tt KL-OSM} rely on the greedy algorithm (Algorithm~\ref{alg:greedy}) to solve the linear maximization for determing the action to be pulled. The time complexity of the greedy algorithm is upper bounded by $\bigO(K(\log K+\mathcal{T}_{\mathrm{member}}))$ time and lower bounded by $\Omega(K)$. 
\cite{perrault2019exploiting} showed that the sampling rule of many combinatorial semi-bandit algorithms is a maximization problem over a summation of a linear function and a submodular function, and proposed two efficient algorithms for matroid semi-bandits: One is based on local search and the other is a greedy algorithm. Both have per-round time complexity at least $\Omega(KD)$.
In contrast, our {\tt FasterCUCB} is the first matroid semi-bandit algorithm with per-round time complexity sublinear in $K$ for many classes of matroids. 

\paragraph{Combinatorial semi-bandits.} 
Here, we review works that focus on the standard setting of stochastic combinatorial semi-bandits. These consider a linear reward function and any action sets $\decls$, where linear maximization $\max_{\bx\in\decls}\inner{\bx}{\bv}$ for any $\bv\in\reals^K$ can be solved in time polynomial in $K$.
We omit the discussion on works that focus on a specific action set \cite{chowdhury2023combinatorial}, with additional structural assumptions on the rewards \cite{wen2015efficient,perrault2020covariance}, or with a different reward function \cite{papadigenopoulos2021recurrent}.
\citet{perrault2020statistical} showed that {\tt CTS} has a regret bound of $\bigO(\frac{K\log^2 D\log T}{\gap{}})$ for mutually independent gaussian rewards
and a regret bound of $\bigO(\frac{KD\log^2 D\log T}{\gap{}})$ for correlated gaussian rewards. \citet{perrault2022combinatorial}
sharpen the regret bound of {\tt CTS} for the case of mutually independent gaussian rewards to be $\bigO(\frac{K\log D\log T}{\gap{}})$. The per-round time complexity of {\tt CTS} is at least $\Omega(K)$ due to sampling from the posterior distributions. 
\citet{degenne2016combinatorial} showed that {\tt ESCB2}  has regret bound of $\bigO(\frac{K\log^2D\log T}{\gap{}})$ for independent subgaussian rewards, but its sampling rule 
is NP-hard \cite{atamturk2017maximizing} to optimize.
\citet{cuvelier2021statistically} proposed {\tt AESCB} that approximates {\tt ESCB2} with per-round time complexity of $\bigO(KD\log^3K\,\poly{\log T})$ while maintaining the same regret bound. Their technique is based on rounding and budgeted-linear maximization.
{\tt OSSB} \cite{combes2017minimal} is an asymptotically instance-specifically optimal algorithm for general structured bandits, including combinatorial semi-bandits, but at each round, it requires to solve a semi-infinite linear program \cite{graves1997asymptotically}. 
\citet{cuvelier2021asymptotically} developed a method that runs in time polynomial in $K$ to solve the semi-infinite linear program for Gaussian rewards. They managed to maintain {\tt OSSB}’s asymptotic optimality for $m$-sets, but not for spanning trees and bipartite matchings.
\citet{ito2021hybrid} and \citet{tsuchiya2023further} proposed algorithms based on the optimistic FTRL framework that achieve $\bigO(\frac{KD\log T}{\gap{}})$ regret in the stochastic setting and $\bigO(\sqrt{KDT\log T})$ in the adversarial setting.
At each round, the proposed algorithms first use FTRL rule to obtain a vector $\ba(t)$ in the convex hull of $\decls$ and then sample an action $\bx(t)$ based on $\ba(t)$. \citet{tsuchiya2023further} mentioned that the computational efficiency of the sampling step has long been a problem in semi-bandits using the optimistic FTRL framework.

\paragraph{Sublinear-time linear bandits.}
Several works \cite{jun2017scalable,yang2022linear} focusing on making per-round complexity of linear bandits sublinear in the number of arms. 
Maximum Inner Product Search (MIPS) is the primary tool used to design such algorithms.
For $N$ arms in $\reals^d$, {\tt Q-GLOC} \cite{jun2017scalable} achieves a high-probability regret bound of $\tilde{\bigO}(d^{\frac{5}{4}}\sqrt{T})$ and per-round time complexity of $\tilde{\bigO}(d^2N^{\rho}\log N)$ for some $\rho\in(0,1)$,
where $\tilde{\bigO}$ hides polylogarithmic factors in $T$ and $d$.
\citet{yang2022linear} considered the setting with arms addition (resp. addition and deletion), and proposed {\tt Sub-Elim} (resp. {\tt Sub-TS}), which has a high-probability regret bound of $\tilde{\bigO}(d\sqrt{T})$ (resp. $\tilde{\bigO}(d^{\frac{3}{2}}\sqrt{T})$) and per-round time complexity of $N^{1-\Theta(\frac{1}{T^2\log^2 T})}$ (resp. $N^{1-\Theta(\frac{1}{T})}$).
These results are applicable to our setting with $d=K$ and $N=\abs{\decls}$.
{\tt Q-GLOC} \cite{jun2017scalable} applied to our setting has regret bound of $\tilde{O}(K^{\frac{5}{4}}\sqrt{T})$ and per-round time compleixty $\tilde{\bigO}(K^2\abs{\decls}^{\rho})$. {\tt Sub-Elim} (resp. {\tt Sub-TS}) \citet{yang2022linear} applied to our setting has regret bound of $\tilde{\bigO}(K\sqrt{T})$ (resp. $\tilde{\bigO}(K^{\frac{3}{2}}\sqrt{T})$) and has per-round time complexity of $\abs{\decls}^{1-\Theta(\frac{1}{T^2\log T})}$ (resp. $\abs{\decls}^{1-\Theta(\frac{1}{T})}$). These results have worse regret bounds than what we have obtained, and their per-round time complexity can be exponential in $K$.

%% file: app_membership.tex
\section{Membership Oracles for Different Matroids}\label{sec:membership}
In this section, we discuss $\mathcal{T}_{\mathrm{member}}$ for the matroids shown in Section~\ref{sec:pre}.
\begin{itemize}
    \item For uniform matroid, the membership oracle is given $\bx\in\icls$ and $k\in[K]\backslash\supp{\bx}$, and has to check whether $\abs{\supp{\bx+\be_k}}\leq D$. Suppose the number $n=\abs{\supp{\bx}}$ is maintained. Then, it takes $\bigO(1)$ time to check whether $n+1 \leq D$, and hence $\mathcal{T}_{\mathrm{member}}=\bigO(1)$.
    
    \item For partition matroids, the membership oracle is given $\bx\in\icls$ and $k\in[K]\backslash\supp{\bx}$, and has to check whether $\abs{\supp{\bx+\be_k}\cap S_i}\leq 1$ for all $i\in[D]$.
    Suppose there is an integer array $A$ of size $K$ such that $j\in S_{A[j]}$ for each $j\in[K]$, and suppose there is an integer array $B$ of size $D$ such that $B[i]=\sum_{j\in\supp{\bx}}\ind{j\in S_i}$ for each $i\in[D]$.
    Then, to decide whether whether $\bx+\be_k\in\icls$, it only requires to check whether $B[A[k]]+1\leq 1$.
    This can be implemented in $\bigO(1)$ time, and thus $\mathcal{T}_{\mathrm{member}}=\bigO(1)$.
    
    \item For graphical matroids, the membership oracle has to detect if there is a cycle. Using the union-find data structure, whether $\supp{\bx}\cup\{k\}$ has a cycle can be detected in $\bigO(\log K)$ time, so we have $\mathcal{T}_{\mathrm{member}}=\bigO(\log K)$. Refer to Section~4.6 in \cite{kleinberg2006algorithm} for more detailed explanation.
    \item For transversal matroids, there is little discussion about its membership oracle. 
    Here we present an implementation to answer a query $(\bx,k)$ about whether $\bx+\be_k\in\icls$, where $\bx\in\icls$ and $k\in[K]\backslash\supp{\bx}$.
    Suppose a maximum matching $M$ on $\supp{\bx}\cup V$ is maintained. 
    Then, answering whether $\bx+\be_k\in\icls$ is equivalent to checking whether an augmenting path on $\supp{\bx+\be_k}\cup V$ from $M$ can be found. Finding a augmentation path can be done by a breadth-first search (BFS) starting from $k$ (see Section 17.2 in \cite{schrijver2003combinatorial}), and it takes $\bigO(DK)$ time because there are at most $K$ leaves in the BFS tree and the length of the path from $k$ to each leaf is at most $2D$. Thus, we have $\mathcal{T}_{\mathrm{member}}=\bigO(DK)$.
\end{itemize}

%% file: app_matroids.tex
\section{Omitted Proofs in \cref{sec:dynamic}}
\label{app:dynamic}

\begin{figure}
    \centering
    \includegraphics[width=0.6\linewidth]{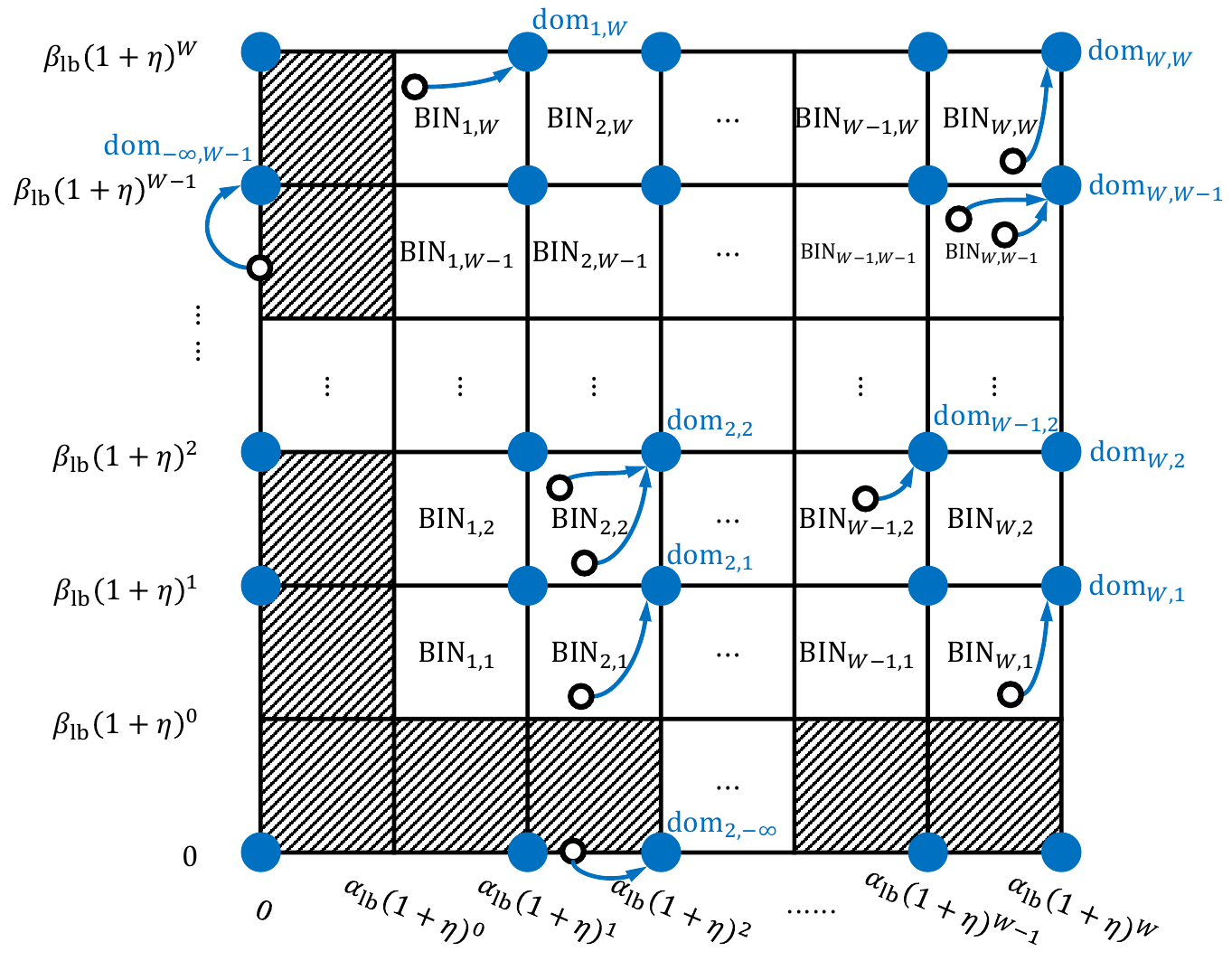}
    \caption{Illustration of feature rounding.
    There are $|\mathbb{W}|^2$ bins, and
    features are assumed not to be in (the interior of) the shaded area.
    Each feature $\bd{f}_k$ is rounded to its dominating point $\dom(\bd{f}_k)$,
    which is specified by a curved arrow.
    }
    \label{fig:rounding}
\end{figure}

\begin{figure}
    \centering
    \includegraphics[width=0.5\linewidth]{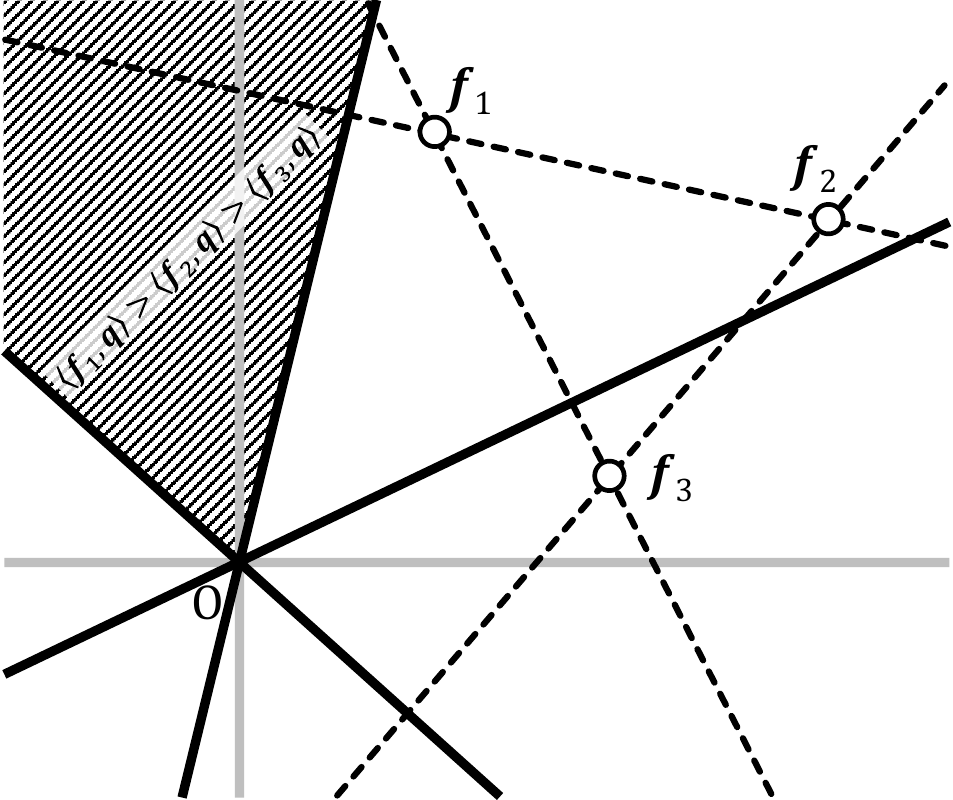}
    \caption{Illustration of characterization of representable permutations.
    There are three features $\bd{f}_1, \bd{f}_2, \bd{f}_3$ on $\reals^2$.
    Each dashed line denotes 
    $\overleftrightarrow{\bd{f}_{i} \bd{f}_{j}}$ for some $i \neq j$;
    each black bold line is orthogonal to some dashed line and intersects the origin.
    Such black bold lines generate six regions, each corresponding to a distinct permutation. 
    For example, for any query $\bd{q}$ in the hatched area, it holds that
    $\langle \bd{f}_1, \bd{q} \rangle > \langle \bd{f}_2, \bd{q} \rangle > \langle \bd{f}_3, \bd{q} \rangle$; i.e.,
    $\bd{q}$ represents a permutation $\pi$ such that
    $(\pi(1), \pi(2), \pi(3)) = (1,2,3)$.
    }
    \label{fig:representable}
\end{figure}

\begin{proof}[Proof of \cref{lem:dynamic:approx}]
By Eq.~\eqref{eq:dynamic:approx} and the optimality of $\bd{x}_\dom^*$,
for any base $\bd{x}$,
we have
\begin{align*}
& \sum_{k \in \supp{\bd{x}_\dom^*}} \langle \bd{f}_k, \bd{q} \rangle \\
& > \frac{1}{1+\err} \cdot
\sum_{k \in \supp{\bd{x}_\dom^*}} \langle \dom(\bd{f}_k), \bd{q} \rangle
& \text{(by Eq.~\eqref{eq:dynamic:approx})} \\
& \geq \frac{1}{(1+\err)^2} \cdot
\sum_{k \in \supp{\bd{x}}} \langle \dom(\bd{f}_k), \bd{q} \rangle
& \text{(by optimality of } \bd{x}_\dom^* \text{)} \\
& \geq \frac{1}{(1+\err)^2} \cdot
\sum_{k \in \supp{\bd{x}}} \langle \bd{f}_k, \bd{q} \rangle
& \text{(by Eq.~\eqref{eq:dynamic:approx})} \\
& \geq \frac{1}{1+\epsilon} \cdot
\sum_{k \in \supp{\bd{x}}} \langle \bd{f}_k, \bd{q} \rangle.
& \text{(as }(1+\err)^2 \leq 1+\epsilon\text{)}\mbox{\qedhere}
\end{align*}
\end{proof}

\begin{proof}[Proof of \cref{lem:dynamic:unique-base}]
The proof follows from the uniqueness of the maximum-weight base in the case of distinct weights;
see, e.g., \cite{edmonds1971matroids}.
\end{proof}

\begin{proof}[Proof of \cref{lem:dynamic:covering}]
For a query $\bd{q} \in \reals^2$, let
$\mathcal{C} \subset \reals^2$
be a cell in arrangement of $\mathcal{L}$ whose
closure contains $\bd{q}$ (which may not be uniquely determined).
Then, there is a permutation $\pi \in \mathfrak{S}_K$ such that
for any vector $\bd{h} \in \mathcal{H} \cap \mathcal{C}$,
we have
$\langle \bd{f}_{\pi(i)}, \bd{h} \rangle > \langle \bd{f}_{\pi(j)}, \bd{h} \rangle$ whenever $i < j$.
Since $\bd{q}$ is in the closure of $\mathcal{C}$,
it holds that
$\langle \bd{f}_{\pi(i)}, \bd{q} \rangle \geq \langle \bd{f}_{\pi(j)}, \bd{q} \rangle$ for any $i < j$,
implying the proof.
\end{proof}

\begin{proof}[Proof of \cref{lem:dynamic:hyperplane-arrangement}]
Since each cell in arrangement of $\mathcal{L}$ is 
a polyhedral cone generated by two lines of $\mathcal{L}$
that does not contain any other lines of $\mathcal{L}$,
there are only $\bigO(K^2)$ cells, and
each of their internal points can be found by using \cref{alg:dynamic:hitting-set}, as desired.

\end{proof}
\begin{algorithm}[h!]
\caption{\textbf{\textsc{Generate-Hitting-Set}}.}
\label{alg:dynamic:hitting-set}
\footnotesize
    \KwIn{
        $K$ distinct features $(\bd{f}_k)_{k \in [K]}$.
    }
    let $\Theta \leftarrow \emptyset$;
    
    \For{all $k \neq k'$}{
        let $L$ be a unique line that is orthogonal to line $\overleftrightarrow{\bd{f}_k \bd{f}_{k'}}$ and
        intersects $\bd{0}$;
        
        add the angle $\theta$ of $L$ and $-\theta$ to $\Theta$;
    }
    let $\mathcal{H} \leftarrow \emptyset$;
    
    \For{all neighboring (but distinct) $\theta_1$ and $\theta_2$ in $\Theta$}{
        let $\bd{h} \triangleq \left(\cos(\frac{\theta_1+\theta_2}{2}), \sin(\frac{\theta_1+\theta_2}{2})\right)$
        be an internal point of a polyhedral cone generated by two half-lines whose angles are $\theta_1$ and $\theta_2$;
        
        add $\bd{h}$ to $\mathcal{H}$;
    }
    return $\mathcal{H}$;
\end{algorithm}

\begin{proof}[Proof of \cref{thm:dynamic}]
The correctness of \textbf{\textsc{Find-Base}} is shown first.
Given a query $\bd{q} \in \reals_+^2$,
\cref{alg:dynamic:base} finds $\bd{h} \in \mathcal{H}$ such that
$\langle \bd{f}_k, \bd{h} \rangle > \langle \bd{f}_{k'}, \bd{h} \rangle$
implies 
$\langle \bd{f}_k, \bd{q} \rangle \geq \langle \bd{f}_{k'}, \bd{q} \rangle$
due to \cref{lem:dynamic:covering}.
Calling $\mathcal{A}_{\bd{h}}$ finds
a $(1+\err)$-approximate maximum-weight base $\bd{x}^\circ$ of $\matcls$
with arm $k$'s weight defined as $\langle \dom(\bd{f}_k), \bd{h} \rangle$.
Since a total order over $[K]$ induced by arm weights $\langle \dom(\bd{f}_k), \bd{h} \rangle$ is consistent with
that induced by arm weights $\langle \dom(\bd{f}_k), \bd{q} \rangle$,
by \cref{lem:dynamic:unique-base},
$\bd{x}^\circ$ is also a $(1+\err)$-approximate maximum-weight base of $\matcls$
with arm $k$'s weight defined as $\langle \dom(\bd{f}_k), \bd{q} \rangle$.
By \cref{lem:dynamic:approx},
\begin{align}
    \sum_{k \in \supp{\bd{x}^\circ}} \langle \bd{f}_k, \bd{q} \rangle \geq 
    \frac{1}{1+\epsilon} \cdot
    \sum_{k \in \supp{\bd{x}}} \langle \bd{f}_k, \bd{q} \rangle,
\end{align}
for any base $\bd{x}$ of $\matcls$; namely,
$\bd{x}^\circ$ is a $(1+\epsilon)$-approximate maximum-weight base of $\matcls$
with arm $k$'s weight defined as
$\langle \bd{f}_k, \bd{q} \rangle$,
completing the correctness of \textbf{\textsc{Find-Base}}.

Subsequently, we bound the time complexity of each subroutine as follows.
\begin{description}
\item[\textbf{\textsc{Initialize}}:]
Construction of $\bin{q,r}$ and $\dom_{q,r}$ for all $q,r \in \mathbb{W}$ completes in $\bigO(K + W^2)$ time.
Then, a hitting set $\mathcal{H}$ for $\dom_{q,r}$'s and $\frac{1}{1+\err}\cdot \dom_{q,r}$'s
of size $|\mathcal{H}| = \bigO(W^4)$ can be constructed in $\poly{W}$ time due to \cref{lem:dynamic:hyperplane-arrangement}.
There will be $|\mathcal{H}|$ instances of algorithm $\mathcal{A}$
(with different arm weights),
creating which takes $\bigO(W^4 \cdot \mathcal{T}_{\mathrm{init}}(\mathcal{A}; \err))$ time.

\item[\textbf{\textsc{Find-Base}}:]
Checking whether
each $\bd{h} \in \mathcal{H}$ and query $\bd{q} \in \reals_+^2$
belong to (the closure of) the same cell in arrangement of $\mathcal{V}$
can be done in $\bigO(W^2)$ time by comparing the induced total orders.
By brute-force search, a desired $\bd{h}$ can be found in $\bigO(W^6)$ time.
Since calling $\mathcal{A}_{\bd{h}}$ requires $\bigO(D)$ time,
the entire time complexity is bounded by $\bigO(\poly{W} + D)$.

\item[\textbf{\textsc{Update-Feature}}:]
For $|\mathcal{H}|$ instances of $\mathcal{A}$,
a single arm's weight would be changed,
each of which runs in $\mathcal{T}_{\mathrm{update}}(\mathcal{A}; \err)$ time.
\end{description}

Observe finally that 
\begin{align}
\begin{aligned}
    W & = \max\left\{
    \left\lceil \log_{1+\err}\left(\frac{\alpha_\ub}{\alpha_\lb}\right) \right\rceil,
    \left\lceil \log_{1+\err}\left(\frac{\beta_\ub}{\beta_\lb}\right) \right\rceil
    \right\} \\
    & = \bigO\left(
        \frac{\log(\frac{\alpha_\ub}{\alpha_\lb}) + \log(\frac{\beta_\ub}{\beta_\lb})}{\log (1+\err)}
    \right) \\
    & = \bigO\left(
    \err^{-1} \cdot \log\left(\frac{\alpha_\ub}{\alpha_\lb} \cdot \frac{\beta_\ub}{\beta_\lb}\right)
    \right) \\
    & = \bigO\left(
    \epsilon^{-1} \cdot \log\left(\frac{\alpha_\ub}{\alpha_\lb} \cdot \frac{\beta_\ub}{\beta_\lb}\right)
    \right),
\end{aligned}
\end{align}
where we used the fact that $\frac{1}{\log(1+\err)} < \frac{1}{\err}$ when $\err \in (0,1)$,
completing the proof.
\end{proof}

\newpage
\section{Proofs Related to Regret Analysis}
\subsection{Proofs Related to the Bijection $g_t$}
\begin{replemma}{lem:gt-property}\label{proof:gt-property}
{\it
There exists a bijection $g_t:\supp{\ist}\to\supp{\bx(t)}$ such that (i) $g_t(j)=j$ for $j\in\supp{\ist}\cap\supp{\bx(t)}$; (ii) for any $j\in\supp{\ist}\backslash\supp{\bx(t)}$,
\[
x_{g_t(j)}(t)=1
\implies
\inner{\dom(\bdf_{g_t(j)})}{\bh}\geq\frac{\inner{\dom(\bdf_j)}{\bh}}{1+\frac{1}{3\log^mT}}.
\]
}
\end{replemma}
\begin{pf}
Let $\eta\triangleq\frac{1}{3\log^mT}$.
The proof is inspired by Section~4.2 in \cite{kveton2014matroid}, and several changes are made to deal with the usage of the dynamic$(1+\eta)$-approximate maximum-weight basis algorithm in the \textbf{\textsc{Find-base}} procedure.

Let $\xi_t:[D]\to\supp{\bx(t)}$ be the ordering such that $\xi_t(i)$'s arm weight $\inner{\dom(\bdf_{\xi_t(i)})}{\bh}$ is the $i$-th largest, where $\bh\in\mathcal{H}$ lies in the same cell as the query $\bq=(1,\lambda_t)$ when invoking \textbf{\textsc{Find-base}} procedure.

\underline{Explicit construction of $g_t$}: 
We define 
\[
g_t(j)=\xi_t(\pi_t^{-1}(j)), \forall j\in\supp{\ist},
\]
where the function $\pi_t:[D]\to\supp{\ist}$ is a bijection such that the following hold: 
\begin{enumerate}
    \item[(i)] $\sum_{i=1}^{k-1}\be_{\xi_t(i)}+\be_{\pi_t(k)}\in\icls$ for all $k\in[D]$
    \item[(ii)] $\pi_t(k)=\xi_t(k)$ if $\xi_t(k)\in\supp{\ist}\cap\supp{\bx(t)}$
\end{enumerate}
The existence of $\pi_t$ is proved in Lemma~\ref{lem:gt-exist} and also by Lemma~1 of \cite{kveton2014matroid}.

\underline{Show (i) $g_t(j)=j$ for $j\in\supp{\ist}\cap\supp{\bx(t)}$}:
Fix any $j\in\supp{\ist}\cap\supp{\bx(t)}$.
From the definition of $\pi_t$, we have $\pi_t(j)=\xi_t(j)$ and hence $g_t(j)=\xi_t(\pi_t^{-1}(j))=\xi_t(\xi_t^{-1}(j))=j$.

\underline{Show (ii) $x_{g_t(j)}(t)=1\implies\inner{\dom(\bdf_{g_t(j)})}{\bh}\geq\frac{\inner{\dom(\bdf_j)}{\bh}}{1+\eta}$}:
Fix any $j\in\supp{\ist}\backslash\supp{\bx}$.
Let $k=\pi_t^{-1}(j)$.
Observe that the bijection $\pi_t$ captures the situation that: The algorithm can choose $\pi_t(k)\in\supp{\ist}$ as the $k$-th element but instead chooses $\xi_t(k)\in\supp{\bx(t)}$. 
By the procedure of Algorithm~\ref{alg:dynamic:base} and Assumption~\ref{asm:dynamic:algorithm}, this happens when 
\[
\inner{\dom(\bdf_{\xi_t(k)})}{\bh}\geq\frac{1}{1+\eta}\inner{\dom(\bdf_{\pi_t(k)})}{\bh},
\]
and replacing $k=\pi_t^{-1}(j)$ completes the proof. 
\end{pf}
\medskip
\begin{lemma}\label{lem:gt-exist}
Let $\bx,\ist\in\decls$, and $\xi:[D]\to\supp{\bx}$ be an arbitrary bijection.
There exists a bijection $\pi:[D]\to\supp{\ist}$ such that $\sum_{i=1}^{k-1}\be_{\xi(i)}+\be_{\pi(k)}\in\icls$ for all $k\in[D]$.
\end{lemma}
\begin{pf}
This lemma is equivalent to Lemma~1 of \cite{kveton2014matroid}. For reader's convenience, we provide a proof here. 

For $k=D$, consider $\sum_{i=1}^{D-1}\be_{\xi(i)}\in\icls$ (due to hereditary property), and $\ist\in\icls$.
As the former has $D-1$ element while the latter has $D$ elements, by augmentation property, there exists $\pi(D)\in\supp{\ist}$ such that $\sum_{i=1}^{D-1}\be_{\xi(i)}+\be_{\pi(D)}\in\icls$.
For the case when $\xi(D)\in\supp{\ist}\cap\supp{\bx}$, we set $\pi(D)=\xi(D)$.

The proof is completed by repeating the following process for $k=D-1,\cdots,1$.
As $\sum_{i=1}^{k-1}\be_{\xi(i)}\in\icls$ (due to hereditary property) has $k-1$ elements, and $\ist-\sum_{i=k+1}^D\be_{\pi(i)}\in\icls$ (due to hereditary property) has $k$ elements, by augmentation property, there exists $\pi(k)$ such that $\sum_{i=1}^{k-1}\be_{\xi(i)}+\be_{\pi(k)}\in\icls$.
If $\xi(k)\in\supp{\ist}\cap\supp{\bx}$, we set $\pi(k)=\xi(k)$.
\end{pf}

\subsection{Lemmas Related to Regret Analysis of Algorithm~\ref{alg:faster-cucb}}
In this section, we fix a best action $\ist\in\argmax_{\bx\in\decls}\inner{\bmu}{\bx}$ and define $\gap{j,k}\triangleq\mu_j-\mu_k$. 
Let $\{\overline{j}\}_{j=1}^D$ be the permutation of $\supp{\ist}$ such that $\mu_{\overline{1}}\geq\cdots\geq\mu_{\overline{D}}$. Define $d_k\triangleq\max\{j\in\supp{\ist}:\gap{\overline{j},k}>0\}$ and $\gap{\min}\triangleq\min_{k\notin\supp{\ist}}\gap{\overline{d_k},k}$.
\begin{replemma}{lem:regret-T0}\label{proof:regret-T0}
{\it
    Let $\epsilon<\frac{\gap{\min}}{b}$. Then, for any $i\in\supp{\ist}$ and any $j\notin\supp{\ist}$, 
    \[
    \mu_i-\mu_j>0 \implies \frac{\mu_i}{1+\epsilon}-\mu_j>0.
    \]
}
\end{replemma}
\begin{pf}
    Fix $i\in\supp{\ist}$ and $j\notin\supp{\ist}$ such that $\mu_i-\mu_j>0$.
    We want the following to hold: 
    \[
        \frac{\mu_i}{1+\epsilon}-\mu_j>0
        \Longleftrightarrow
        \mu_i-(1+\epsilon)\mu_j>0\\
        \Longleftrightarrow
        \mu_i-\mu_j>\epsilon\mu_j.
    \]
    As $\mu_i-\mu_j>\epsilon\mu_j$ must hold for all such $i$ and $j$, taking the minimum over all possible $i$ and $j$ on the left-hand side, and use the fact that $\mu_j\leq b$ for all $j$ on the right-hand side, we derive 
    \[
    \frac{\gap{\min}}{b} > \epsilon
    \]
    is the condition on $\epsilon$ to ensure $\mu_i-\mu_j>0 \implies \frac{\mu_i}{1+\epsilon}-\mu_j>0$ holds for all $i$ and $j$.
\end{pf}

\medskip
\begin{replemma}{lem:regret-anytime-part1}\label{proof:regret-anytime-part1}
{\it
Let $k\notin\supp{\ist}$ and $j\in[d_k]$. For $T>T_0$,
\[
\sum_{j=1}^{d_k}\gap{\overline{j},k}(I)_{\overline{j},k} 
\leq 
\sum_{j=1}^{d_k}\gap{\overline{j},k}T_0
+\frac{12(b-a)^2\gap{\overline{d_k},k}\log T}{(\frac{\mu_{\overline{d_k}}}{1+\log^{-m}T}-\mu_k)^2}.
\]
}
\end{replemma}
\begin{pf}
Recall $(I)_{\overline{j},k}=\sum_{t=1}^T\expectation{}{\ind{g_t(\overline{j})=k, N_k(t)\leq n_{\overline{j},k}}}$, where $n_{j,k}=\max\left\{\frac{6(b-a)^2\log T}{(\frac{\mu_j}{1+\log^{-m}T}-\mu_k)^2},T_0\right\}$.

First, we claim that: for any $\{a_j\}_{j=1}^{d_k}$ with $a_1\geq\cdots\geq a_{d_k}\geq 0$, 
\begin{equation}\label{eq:regret-anytime-part1-0}    \sum_{j=1}^{d_k}a_j(I)_{\overline{j},k} \leq a_1n_{\overline{1},k}+\sum_{j=2}^{d_k}a_j(n_{\overline{j},k}-n_{\overline{j-1},k}).
\end{equation}
\underline{Show Eq.~\eqref{eq:regret-anytime-part1-0}}:
We show by induction. 
For the base case, we have 
\begin{equation}\label{eq:regret-anytime-part1-1}
a_1(I)_{\overline{1},k}+a_2(I)_{\overline{2},k} 
\leq
a_1n_{\overline{1},k} + a_2(n_{\overline{2},k}-n_{\overline{1},k}).
\end{equation}
Eq.~\eqref{eq:regret-anytime-part1-1} is derived as follows.
Since $a_1,a_2\geq 0$ and 
\[
(I)_{\overline{1},k}+(I)_{\overline{2},k}
=\sum_{t=1}^T\expectation{}{\ind{g_t(\overline{1})=k, N_k(t)\leq n_{\overline{1},k}}+\ind{g_t(\overline{2})=k, N_k(t)\leq n_{\overline{2},k}}} 
\leq \max\{n_{\overline{1},k}, n_{\overline{2},k}\}=n_{\overline{2},k},
\]
therefore we can bound $(I)_{\overline{2},k}$ as $(I)_{\overline{2},k}\leq n_{\overline{2},k}-(I)_{\overline{1},k}$, yields that: 
\[
a_1(I)_{\overline{1},k}+a_2(I)_{\overline{2},k}
\leq (a_1-a_2)(I)_{\overline{1},k}+\gap{\overline{2},k}n_{\overline{2},k}.
\]
Then, since $a_1\geq a_2$ and  $(I)_{\overline{1},k}\leq n_{\overline{1},k}$, we derive 
\[
a_1(I)_{\overline{1},k}+a_2(I)_{\overline{2},k}
\leq (a_1-a_2)n_{\overline{1},k}+a_2n_{\overline{2},k},
\]
which shows Eq.~\eqref{eq:regret-anytime-part1-1}.
Now, assume for any $\{b_j\}_{j=1}^{\ell}$ with $b_1\geq\cdots\geq b_{\ell}\geq 0$, the following 
\begin{equation}\label{eq:regret-anytime-part1-2}
\sum_{j=1}^{\ell}b_j(I)_{\overline{j},k}\leq b_1n_{\overline{1},k}+\sum_{j=2}^{\ell}b_j(n_{\overline{j},k}-n_{\overline{j-1},k})
\end{equation}
holds for $\ell<d_k$.

Fix any $\{a_j\}_{j=1}^{\ell+1}$ with $a_1\geq\cdots\geq a_{\ell+1}\geq 0$.
Consider $\sum_{j=1}^{\ell+1}a_j(I)_{\overline{j},k}$.
Since $a_j\geq 0$ for all $j\in[\ell+1]$ and 
\[
\sum_{j=1}^{\ell+1}(I)_{\overline{j},k}
=\sum_{t=1}^T\expectation{}{\sum_{j=1}^{\ell+1}\ind{g_t(\overline{j})=k, N_k(t)\leq n_{\overline{1},k}}}
\leq \max_{j\in[\ell+1]}n_{\overline{j},k}=n_{\overline{\ell+1},k},
\]
we can bound $(I)_{\overline{\ell+1},k}$ as $(I)_{\overline{\ell+1},k}\leq n_{\overline{\ell+1},k}-\sum_{j=1}^{\ell}(I)_{\overline{j},k}$, which results in:
\[
\sum_{j=1}^{\ell+1}a_j(I)_{\overline{j},k}
\leq \sum_{j=1}^{\ell}(a_j-a_{\ell+1})(I)_{\overline{j},k} + a_{\ell+1}n_{\overline{\ell+1},k}.
\]
Since $a_1-a_{\ell+1}\geq \cdots\geq a_{\ell}-a_{\ell+1}\geq 0$, using inductive hypothesis Eq.~\eqref{eq:regret-anytime-part1-2} with $b_j=a_j-a_{\ell+1}$ for all $j\in[\ell]$, we get
\begin{align*}
\sum_{j=1}^{\ell+1}a_j(I)_{\overline{j},k}
& \leq 
(a_1-a_{\ell+1})n_{\overline{1},k} + \sum_{j=2}^{\ell}(a_j-a_{\ell+1})(n_{\overline{j},k}-n_{\overline{j-1},k}) 
+a_{\ell+1}n_{\overline{\ell+1},k}. \\
& =a_1n_{\overline{1},k} +\sum_{j=2}^{\ell}a_j(n_{\overline{j},k}-n_{\overline{j-1},k})
-a_{\ell+1}\left(n_{\overline{1},k}
+\sum_{j=2}^{\ell}(n_{\overline{j},k}-n_{\overline{j-1},k})
-n_{\overline{\ell+1},k}\right)\\
& = a_1n_{\overline{1},k} +\sum_{j=2}^{\ell}a_j(n_{\overline{j},k}-n_{\overline{j-1},k})
+a_{\ell+1}(n_{\overline{\ell+1},k}-n_{\overline{\ell},k})\\
& = a_1n_{\overline{1},k} +\sum_{j=2}^{\ell+1}a_j(n_{\overline{j},k}-n_{\overline{j-1},k}).
\end{align*}
Thus, Eq.~\eqref{eq:regret-anytime-part1-0} is proved by induction.

Define $\epsilon\triangleq\log^{-m}T$ and $\gap{j,k}(\epsilon)\triangleq\frac{\mu_j}{1+\epsilon}-\mu_k$. 
Using Eq.~\eqref{eq:regret-anytime-part1-0} with $a_j=\gap{\overline{j},k}$ for $j\in[d_k]$ and recalling $n_{j,k}\triangleq\max\left\{\frac{6(b-a)^2\log T}{\gap{j,k}(\epsilon)^2},T_0\right\}$, we have 
\begin{align*}
\sum_{j=1}^{d_k}\gap{\overline{j},k}(I)_{\overline{j},k}
& \leq \gap{\overline{1},k}n_{\overline{1},k} + \sum_{j=2}^{d_k}\gap{\overline{j},k}(n_{\overline{j},k}-n_{\overline{j-1},k})\\
& \leq \sum_{j=1}^{d_k}\gap{\overline{j},k}T_0 + 6(b-a)^2\log T\left(\frac{\gap{\overline{1},k}}{\gap{\overline{1},k}(\epsilon)^2} + \sum_{j=2}^{d_k}\gap{\overline{j},k}\left(
    \frac{1}{\gap{\overline{j},k}(\epsilon)^2}
    -\frac{1}{\gap{\overline{j-1},k}(\epsilon)^2}
\right)\right).
\end{align*}
We upper bound the last term by: 
\begin{align*}
\frac{\gap{\overline{1},k}}{\gap{\overline{1},k}(\epsilon)^2} + \sum_{j=2}^{d_k}\gap{\overline{j},k}\left(
    \frac{1}{\gap{\overline{j},k}(\epsilon)^2}
    -\frac{1}{\gap{\overline{j-1},k}(\epsilon)^2}
\right)
& =\sum_{j=1}^{d_k-1}\frac{\gap{\overline{j},k}(\epsilon)-\gap{\overline{j+1},k}(\epsilon)}{(\gap{\overline{j},k}(\epsilon))^2}+\frac{\gap{\overline{d_k},k}}{\gap{\overline{d_k},k}(\epsilon)^2}\\
& \leq \sum_{j=1}^{d_k-1}\frac{\gap{\overline{j},k}(\epsilon)-\gap{\overline{j+1},k}(\epsilon)}{\gap{\overline{j},k}(\epsilon)\gap{\overline{j+1},k}(\epsilon)}
+\frac{\gap{\overline{d_k},k}}{\gap{\overline{d_k},k}(\epsilon)^2}\\
& =
\sum_{j=1}^{d_k-1}\left(\frac{1}{\gap{\overline{j+1},k}(\epsilon)}-\frac{1}{\gap{\overline{j},k}(\epsilon)}\right)
+\frac{\gap{\overline{d_k},k}}{\gap{\overline{d_k},k}(\epsilon)^2}
\leq \frac{2\gap{\overline{d_k},k}}{\gap{\overline{d_k},k}(\epsilon)^2},
\end{align*}
where the first inequality is due to $\gap{\overline{j},k}(\epsilon)\geq\gap{\overline{j+1},k}(\epsilon)$, and the second upperbounds the telescoping series:
\[
\sum_{j=1}^{d_k-1}\left(\frac{1}{\gap{\overline{j+1},k}(\epsilon)}-\frac{1}{\gap{\overline{j},k}(\epsilon)}\right)
=\frac{1}{\gap{\overline{d_k},k}(\epsilon)}-\frac{1}{\gap{\overline{1},k}(\epsilon)}
\leq \frac{1}{\gap{\overline{d_k},k}(\epsilon)}
\]
Hence, we derive an upper bound for the first part relevant to $(I)_{\overline{j},k}$: 
\[
\sum_{j=1}^{d_k}\gap{\overline{j},k}(I)_{\overline{j},k} 
\leq 
\sum_{j=1}^{d_k}\gap{\overline{j},k}T_0
+\frac{12(b-a)^2\gap{\overline{d_k},k}\log T}{\gap{\overline{d_k},k}(\epsilon)^2}.
\]
\end{pf}

\begin{replemma}{lem:regret-part2}\label{proof:regret-part2}
{\it
Let $k\notin\supp{\ist}$ and $j\in[d_k]$. For $T>T_0$,
\[
(II)_{\overline{j},k}
\leq \frac{1}{T}+\frac{\pi^2}{6}.
\]
}
\end{replemma}
\begin{proofof}{Lemma~\ref{lem:regret-part2}}
Let $\epsilon=\frac{1}{\log^mT}$.
Recall 
\[(II)_{\overline{j},k}=\sum_{t=1}^T\expectation{}{\ind{g_t(\overline{j})=k, N_k(t)> n_{\overline{j},k}}},
\]
where $n_{j,k}=\max\left\{\frac{6(b-a)^2\log T}{(\frac{\mu_j}{1+\log^{-m}T}-\mu_k)^2},T_0\right\}$.

First, we claim that
\begin{equation}\tag{\ref{eq:regret-e0}}
g_t(\overline{j})=k
\implies
u_{k}(N_{k}(t-1),T)
\geq \frac{\min_{s<t}u_{\overline{j}}(s,t)}{1+\epsilon},
\end{equation}
where $u_k(s,t)=\tilde{\mu}_k(s)+\sqrt{\frac{1.5(b-a)^2\log t}{s}}$ and $\tilde{\mu}_k(t)=\frac{1}{t}\sum_{s=1}^ty_k(s)$.

\underline{Show Eq.~\eqref{eq:regret-e0}}:
Observe that $g_t(\overline{j})=k$ implies
\[
\left(1+\frac{\epsilon}{3}\right)\inner{\bdf_{k}}{\bq}
\geq \inner{\dom(\bdf_{k})}{\bq}
\geq
\frac{\inner{\dom(\bdf_{\overline{j}})}{\bq}}{1+\frac{\epsilon}{3}}
\geq
\frac{\inner{\bdf_{\overline{j}}}{\bq}}{1+\frac{\epsilon}{3}},
\]
where Eq.~\eqref{eq:dynamic:approx} is used in the first and the last inequality, and the second inequality is due to Lemma~\ref{lem:gt-property} and Corollary~\ref{cor:dynamic:covering-new}.
By $(1+\frac{\epsilon}{3})^2\leq 1+\epsilon$ and expanding $\bdf_k=(\hat{\mu}_k(t-1),\frac{1}{\sqrt{N_k(t-1)}})$ and $\bq=(1,\sqrt{1.5(b-a)^2\log t})$, we have
\[
u_{k}(N_{k}(t-1),t)\\
\geq \frac{u_{\overline{j}}(N_{\overline{j}}(t-1),t)}{1+\epsilon}.
\]
As $\log T>\log t$ and $N_{\overline{j}}(t-1)\in[t-1]$, we further derive 
\begin{align*}
u_{k}(N_{k}(t-1),T) \geq \frac{u_{\overline{j}}(N_{\overline{j}}(t-1),t)}{1+\epsilon}
\geq \frac{\min_{s<t}u_{\overline{j}}(s,t)}{1+\epsilon},
\end{align*}
which shows Eq.~\eqref{eq:regret-e0}.

Second, let $\mathcal{T}_{\overline{j},k}=\{t\in\{n_{\overline{j},k}+1,\cdots,T\}: g_t(\overline{j})=k, N_k(t-1)>n_{\overline{j},k}\}$.
From Eq.~\eqref{eq:regret-e0}, we derive 
\begin{align}
(II)_{\overline{j},k} & =\sum_{t=n_{\overline{j},k}+1}^T\prob{}{g_t(\overline{j})=k, N_k(t-1)>n_{\overline{j},k}} \nonumber\\
& \leq\sum_{t=n_{\overline{j},k}+1}\prob{}{u_{k}(N_k(t-1),T)\geq \frac{\min_{s<t}u_{\overline{j}}(s,t)}{1+\epsilon}\text{ and }t\in\mathcal{T}_{\overline{j},k}
}
\nonumber\\
& \leq \sum_{t=n_{\overline{j},k}+1
}\sum_{s<t}
\prob{}{u_k(N_k(t-1),T) \geq \frac{u_{\overline{j}}(s,t)}{1+\epsilon}\text{ and }t\in\mathcal{T}_{\overline{j},k}}
\label{eq:regret-part2-sketch-e0},
\end{align}
where the last inequality uses union bound.

Third, we now upper bound each term $\prob{}{u_k(N_k(t-1),T) \geq \frac{u_{\overline{j}}(s,t)}{1+\epsilon}\text{ and }t\in\mathcal{T}_{\overline{j},k}}$ in Eq.~\eqref{eq:regret-part2-sketch-e0}.
Remind that  
\[
u_k(N_k(t-1),T) \geq \frac{u_{\overline{j}}(s,t)}{1+\epsilon}
\Longleftrightarrow
\underbrace{\tilde{\mu}_k(N_k(t-1))+\frac{\lambda_T}{\sqrt{N_k(t-1)}}}_{A_t} 
\geq 
\underbrace{\frac{\tilde{\mu}_{\overline{j}}(s)+\frac{\lambda_t}{\sqrt{s}}}{1+\epsilon}}_{B_{t,s}}.
\]
Define the event $\mathcal{E}_{t,s}=\{A_t\geq B_{t,s}\text{ and }t\in\mathcal{T}_{\overline{j},k}\}$.
We will partition the event $\mathcal{E}_{t,s}$ by comparing $A_t$ to $A'_t=\mu_k+\frac{2\lambda_T}{\sqrt{N_k(t-1)}}$ and comparing $B_{t,s}$ to $B'=\frac{\mu_{\overline{j}}}{1+\epsilon}$ as follows:
\begin{itemize}
    \item $\mathcal{E}_{t,s}\cap\left\{A_t \geq A'_t\text{ and }t\in\mathcal{T}_{\overline{j},k}\right\} \subseteq\left\{\tilde{\mu}_k(N_k(t-1)) \geq \mu_k+\frac{\lambda_T}{\sqrt{N_k(t-1)}}\text{ and }t\in\mathcal{T}_{\overline{j},k}\right\}$
    
    \item $\mathcal{E}_{t,s}\cap\{B_{t,s} \leq B'\text{ and }t\in\mathcal{T}_{\overline{j},k}\}\subseteq \left\{\mu_{\overline{j}} \geq \tilde{\mu}_{\overline{j}}(s)+\frac{\lambda_t}{\sqrt{s}}\text{ and }t\in\mathcal{T}_{\overline{j},k}\right\}$
    
    \item $\mathcal{E}_{t,s}\cap\{A_t < A_t'\text{ and }B_{t,s} > B'\text{ and }t\in\mathcal{T}_{\overline{j},k}\}\subseteq
    \left\{
        \mu_k+\frac{2\lambda_T}{\sqrt{N_k(t-1)}}
        >\frac{\mu_{\overline{j}}}{1+\epsilon}
    \text{ and }t\in\mathcal{T}_{\overline{j},k}
    \right\}$.
    The inclusion is because under the event $\mathcal{E}_{t,s}\cap\{A_t < A_t'\text{ and }B_{t,s} > B'\text{ and }t\in\mathcal{T}_{\overline{j},k}\}$, we have  
    \[
    \mu_k+\frac{2\lambda_T}{\sqrt{N_k(t-1)}}=A_t' 
    > A_t \geq B_{t,s} 
    > B'=\frac{\mu_{\overline{j}}}{1+\epsilon},
    \]
    where the first and last inequalities are due to the event $\{A_t < A_t'\text{ and }B_{t,s} > B'\text{ and }t\in\mathcal{T}_{\overline{j},k}\}$, and the second inequality is due to the event $\mathcal{E}_{t,s}=\{A_t\geq B_{t,s}\text{ and }t\in\mathcal{T}_{\overline{j},k}\}$.  
\end{itemize}
Hence, we have the following inclusion:
\begin{align*}
& \{A_t \geq A'_t\text{ and }t\in\mathcal{T}_{\overline{j},k}\}
\cup\{B_{t,s} \leq B'\text{ and }t\in\mathcal{T}_{\overline{j},k}\}
\cup\{A_t < A_t'\text{ and }B_{t,s} > B'\text{ and }t\in\mathcal{T}_{\overline{j},k}\}\\
& =\{t\in\mathcal{T}_{\overline{j},k}\}
\supset
\{A_t\geq B_{t,s}\text{ and }t\in\mathcal{T}_{\overline{j},k}\}
=\mathcal{E}_{t,s}.
\end{align*}
From union bound, 
\begin{align}
\prob{}{\mathcal{E}_{t,s}}
& \leq \prob{}{\{A_t \geq A'_t\text{ and }t\in\mathcal{T}_{\overline{j},k}\}\cap\mathcal{E}_{t,s}} +\prob{}{\{B_{t,s} \leq B'\text{ and }t\in\mathcal{T}_{\overline{j},k}\}\cap\mathcal{E}_{t,s}} \nonumber\\
& \quad +\prob{}{\{A_t < A_t'\text{ and }B_{t,s} > B'\text{ and }t\in\mathcal{T}_{\overline{j},k}\}\cap\mathcal{E}_{t,s}} \nonumber\\
& \leq \prob{}{
    \mu_k+\frac{\lambda_T}{\sqrt{N_k(t-1)}} \leq \tilde{\mu}_k(N_k(t-1))
    \text{ and }t\in\mathcal{T}_{\overline{j},k}
}\nonumber\\
& +\prob{}{\mu_{\overline{j}} \geq \tilde{\mu}_{\overline{j}}(s)+\frac{\lambda_t}{\sqrt{s}}\text{ and }t\in\mathcal{T}_{\overline{j},k}}
+\prob{}{\mu_k+\frac{2\lambda_T}{\sqrt{N_k(t-1)}} > \frac{\mu_{\overline{j}}}{1+\epsilon}\text{ and }t\in\mathcal{T}_{\overline{j},k}}.\label{eq:regret-part2-e1}
\end{align}
In Eq.~\eqref{eq:regret-part2-e1}, recall $\lambda_t=\sqrt{1.5(b-a)^2\log t}$ and observe that the last term  
\[
\prob{}{\mu_k+2\sqrt{\frac{1.5(b-a)^2\log T}{N_k(t-1)}} \geq \frac{\mu_{\overline{j}}}{1+\epsilon}\text{ and }t\in\mathcal{T}_{\overline{j},k}}
\leq 
\prob{}{\mu_k+2\sqrt{\frac{1.5(b-a)^2\log T}{n_{\overline{j},k}+1}} \geq \frac{\mu_{\overline{j}}}{1+\epsilon}}
=0,
\]
where the inequality is because $t\in\mathcal{T}_{\overline{j},k}$ implies $N_k(t-1)\geq n_{\overline{j},k}+1$, and the equality is because 
\[
n_{\overline{j},k}
\geq 
\frac{6(b-a)^2\log T}{(\frac{\mu_{\overline{j}}}{1+\epsilon}-\mu_k)^2}
\implies
\frac{6(b-a)^2\log T}{n_{\overline{j},k}+1} < \left(\frac{\mu_{\overline{j}}}{1+\epsilon}-\mu_k\right)^2
\]
and also we have $\frac{\mu_{\overline{j}}}{1+\epsilon}-\mu_k>0$ which is ensured by Lemma~\ref{lem:regret-T0} as $T>T_0$.
Finally, from Eq.~\eqref{eq:regret-part2-sketch-e0} and Eq.~\eqref{eq:regret-part2-e1}, 
\begin{align*}
(II)_{\overline{j},k} 
& \leq \sum_{t=n_{\overline{j},k}+1}^T\sum_{s<t}
    \prob{}{\tilde{\mu}_k(N_k(t-1)) \geq \mu_k+\sqrt{\frac{1.5(b-a)^2\log T}{N_k(t-1)}}\text{ and }t\in\mathcal{T}_{\overline{j},k}}\\
& \quad +\sum_{t=n_{\overline{j},k}+1}^T\sum_{s<t}
\prob{}{\mu_{\overline{j}} \geq \tilde{\mu}_{\overline{j}}(s)+\sqrt{\frac{1.5(b-a)^2\log t}{s}}\text{ and }t\in\mathcal{T}_{\overline{j},k}}
\\
& \leq \sum_{t=n_{\overline{j},k}+1}^T\sum_{s<t}
    \left(\prob{}{\tilde{\mu}_k(t-1) \geq \mu_k+\sqrt{\frac{1.5(b-a)^2\log T}{t-1}}}
    +\prob{}{\mu_{\overline{j}} \geq \tilde{\mu}_{\overline{j}}(s)+\sqrt{\frac{1.5(b-a)^2\log t}{s}}}\right)\\
& \leq \sum_{t=n_{\overline{j},k}+1}^T\sum_{s<t}\left(e^{-3\log T}+e^{-3\log t}\right),
\end{align*}
where the second inequality is because $\{N_k(t-1)\}_{t\in\mathcal{T}_{\overline{j},k}}$ is strictly increasing (as $N_k(t)=N_k(t-1)+1$ when $g_t(\bar{j})=k$) and thus is a subsequence of $\{n_{\overline{j},k}+1,\cdots,T\}$, and the last inequality is due to an application of Hoeffding's inequality (Lemma~\ref{lem:hoeffding}) with $s=\sqrt{1.5(t-1)(b-a)^2\log T}$ and $n=t-1$ to bound the first term and with $s=\sqrt{1.5s(b-a)^2\log t}$ and $n=s$ to bound the second term. 
The proof is completed by evaluating
\begin{align*}
& \sum_{t=1}^T\sum_{s<t}e^{-3\log T}\leq\sum_{t=1}^T\frac{t}{T^3}
\leq \frac{T(T+1)}{2T^3}
\leq\frac{1}{T},\\
& \sum_{t=1}^T\sum_{s<t}e^{-3\log t}
\leq \sum_{t=1}^{\infty}\frac{t}{t^3}
\leq \sum_{t=1}^{\infty}\frac{1}{t^2} \leq \frac{\pi^2}{6}.
\end{align*}
\end{proofof}

\begin{lemma}[Hoeffding's inequality]\label{lem:hoeffding}
    Let $X_1,\cdots,X_n$ be independent random variables such that $X_i\in[a,b]$ for all $i \in [n]$. Then, for all $s >0$,
    \[
    \prob{}{\sum_{i=1}^n(X_i-\expectation{}{X_i}) \geq s} \leq \exp\left(-\frac{2s^2}{n(b-a)^2}\right).
    \]
\end{lemma}